\documentclass[10pt,twocolumn,letterpaper]{article}

\usepackage{iccv}

\usepackage{times,epsfig,graphics,graphicx,caption,float,subcaption,booktabs,xcolor,multirow,multicol,array,color,ifthen,tabu,colortbl,url,xparse,mathtools,patchcmd,enumitem,amssymb,xspace,nicefrac,microtype,amsmath,amsfonts}

\usepackage[title]{appendix}

\newcommand{\Ll}{l}
\newcommand{\La}{\mathbf{\alpha}}

\newcommand{\ours}{F-Clip}

\newcommand{\dai}[1]{\textcolor[rgb]{1,0,0}{#1}}

\usepackage[pagebackref=true,breaklinks=true,letterpaper=true,colorlinks,bookmarks=false]{hyperref}

\iccvfinalcopy 

\def\iccvPaperID{9710} 

\ificcvfinal\fi

\begin{document}

\title{Fully Convolutional Line Parsing}

\author{
Xili Dai \\
HKUST \\
{\small daixili.cs@gmail.com}
\and
Haigang Gong\\
UESTC\\
{\small hggong@uestc.edu.cn}
\and
Shuai Wu \\
UESTC\\
{\small shuaiwu.ws@gmail.com}
\and
Xiaojun Yuan\\
UESTC\\
{\small xjyuan@uestc.edu.cn}
\and
Yi Ma\\
UC Berkeley\\
{\small yima@eecs.berkeley.edu}
}
\maketitle
\ificcvfinal\thispagestyle{empty}\fi


\begin{abstract}
We present a one-stage \textbf{F}ully \textbf{C}onvolutional \textbf{Li}ne \textbf{P}arsing network (\ours) that detects line segments from images. The proposed network is very simple and flexible with variations that gracefully trade off between speed and accuracy for different applications. \ours{} detects line segments in an end-to-end fashion by predicting each line's center position, length, and angle. We further customize the design of convolution kernels of our fully convolutional network to effectively exploit the statistical priors of the distribution of line angles in real image datasets. We conduct extensive experiments and show that our method achieves a significantly better trade-off between efficiency and accuracy, resulting in a real-time line detector at up to 73 FPS on a single GPU. Such inference speed makes our method readily applicable to real-time tasks without compromising any accuracy of previous methods. Moreover,  when equipped with a performance-improving backbone network, \ours{} is able to significantly outperform all state-of-the-art line detectors on accuracy at a similar or even higher frame rate. In other word, under same inference speed, \ours{} always achieving best accuracy compare with other methods. Source code \textit{\dai{https://github.com/Delay-Xili/F-Clip}}.\footnote{This paper has been accepted by \href{https://doi.org/10.1016/j.neucom.2022.07.026}{Neurocomputing}. This version is same as the Neurocomputing one except the latex template.}
\end{abstract}

\section{Introduction}

A holistic 3D representation aims to model and  reconstruct a scene with high-level geometric primitives/structures such as lines, planes, and layouts~\cite{zhou2020holicity}. Unlike representations based on local features that are usually noisy and incomplete, a holistic counterpart is arguably more compact, robust, and easy to use. This belief has motivated a line of recent works on recognizing geometric structures from image observations~\cite{huang2018learning,zhou2019end,zhou2019learning,liu2018planenet,liu2019planercnn,mousavian2019visual,zou2018layoutnet,zeng2020bundle}.

Among all the geometric primitives mentioned above, lines are arguably the most important and fundamental one. An accurate line detection system is essential for many down-stream vision tasks such as vanishing point detection~\cite{zhou2019neurvps}, camera pose estimation~\cite{elqursh2011line}, camera calibration~\cite{zhou2013line}, stereo matching~\cite{yu2013line}, and even full 3D reconstruction~\cite{denis2008efficient, zhou2019learning}.

\begin{figure}[t]
\begin{center}
  \includegraphics[width=0.695\linewidth]{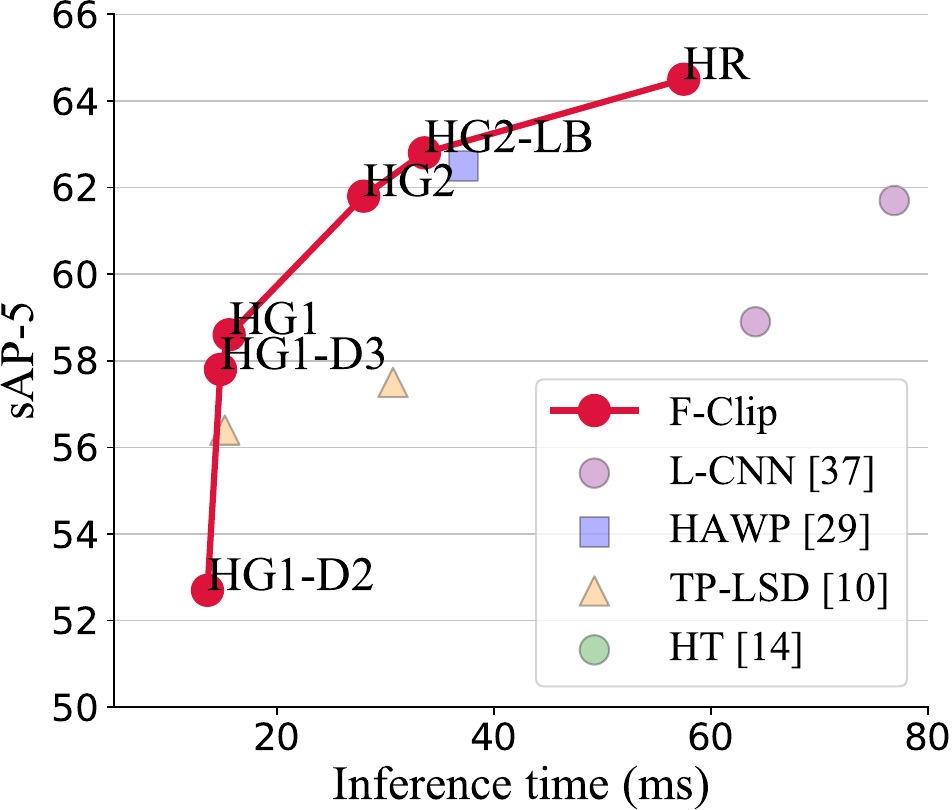}
\end{center}
\vskip -0.15in
\caption{\small 
Speed (ms) versus accuracy (sAP${}^5$) trade-off of state-of-the-art algorithms on the ShanghaiTech wireframe dataset. 
}
\label{fig:teaser}
\end{figure}

Recently, significant progress has been made in the field of line detection due to the introduction of a large-scale dataset~\cite{huang2018learning}, effective learning methods~\cite{zhou2019end}, and the community's continuing effort to develop better algorithms~\cite{xue2019learning,lin2020deep,xue2020holistically}. Nevertheless, most of the existing methods focus primarily on \textit{accuracy}, and their performance drops significantly if modified for \textit{efficiency}. 

In this work, our goal is to develop a flexible algorithm that can achieve the best speed-accuracy trade-off (Figure~\ref{fig:teaser}).

We argue that the unsatisfactory speed-accuracy trade-off of existing methods mostly comes from the nature of the typical ``two-stage'' model design~\cite{zhou2019end,xue2020holistically}. In the first stage, thousands of line candidates are extracted. After that, based on the proposed lines, one extracts the corresponding image features and trains a small sub-network to determine whether each proposed line is correct. This approach has shown to be effective and achieves state-of-the-art accuracy so far. 

However, such a two-stage method sacrifices efficiency since it needs to process a large number of line candidates with an extra sub-network. Structure-wise, such a two-stage model also lacks flexibility in case we need to change the network to achieve a graceful trade-off between speed and accuracy. 

As described in the seminal work~\cite{zhou2019end}, the ``two-stage'' line detection methods are motivated by ``two-stage'' object detection~\cite{girshick2014rich,girshick2015fast,ren2015faster}. Beyond the prevalent two-stage methods, there has been another line of work in object detection known as ``one-stage'' methods. One-stage object detection is done using a dense sliding window, implemented by a fully convolutional network. Such one-stage methods are believed to be more flexible and efficient in certain tasks, and they can achieve more than 200 FPS with decent accuracy~\cite{redmon2016you,liu2016ssd,redmon2017yolo9000,redmon2018yolov3}. Hence, in this work we take on the following question:
\begin{quote}
    \textit{Can we achieve a better speed-accuracy trade-off in line detection by leveraging the successful ideas of one-stage methods in object detection?}
\end{quote}

\vspace{0.1in}
\noindent {\bf Contributions of This Paper.} A key observation that motivates this work is that the extensive line proposal of the first stage used in most existing methods is not entirely necessary. Instead, a line segment can be viewed as an object and can be conveniently represented by its center, length, and angle. Hence, we can formulate the prediction of each of the parameter as a pixel-wise classification/regression problem. To this end, we propose a Fully Convolutional Line Parsing (\ours) network, which realizes the above idea via a fully convolutional network. Besides that, the key contribution of this paper is the achievement of best speed-accuracy trade-off (Figure~\ref{fig:teaser}). In other words, we always get best performance under similar speed compare with other methods.

\begin{figure}[t]
\centering
\includegraphics[width=0.99\linewidth]{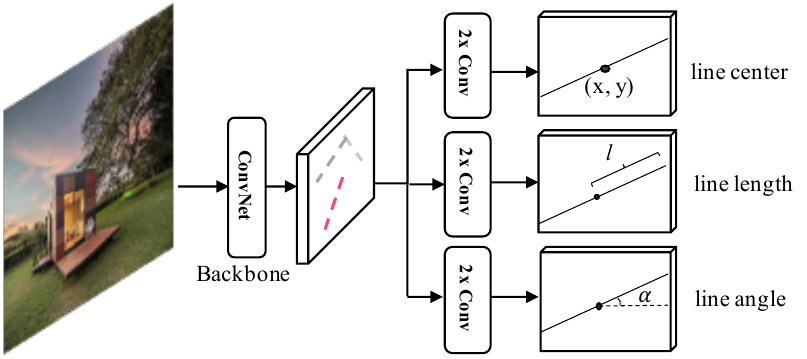}
\caption{\small The overall architecture of the \ours network. Taken an image as input, a backbone convolutional network is followed by three convolution heads that predict the line segment center, length, and angle respectively as the output.}
\label{fig:net}
\end{figure}

\ours{} has a surprisingly simple architecture as illustrated in Figure~\ref{fig:net}. It does not require any heavy network engineering or carefully designed training samplers as in ~\cite{zhou2019end}. To detect a line, our system simply applies a convolutional neural network to extract the image features and uses two additional convolution layers to regress the center, length, and angle score map. Then, for each line center with a high score, we directly output a line segment by associating the length and angle values in the same location. Due to its simple one-stage design, the network is amenable to variations and modifications for different speed-accuracy trade-off, as we will discuss in more detail in Section \ref{sec::method}. 

Through extensive experiments on large real-world image datasets, we will see that the proposed simple method/network achieves a surprisingly good speed-accuracy trade-off. In the latency sensitive setting, \ours{} with a simple hour-glass backbone \cite{newell2016stacked} can achieve 52.7 sAP${}^{5}$ at 73 FPS, nearly 5 times faster than \cite{zhou2019end} at a similar accuracy. Equipped with a performance-improving backbone network~\cite{sun2019deep}, \ours{} can achieve 64.3 sAP${}^{5}$ at 17.4 FPS, better than the state-of-the-art method \cite{huang2020tp}. 

\section{Related Work}\label{sec:prework}

\paragraph{Line Detection} The classical approach for line detection can be dated back to the 70s'. Hough transform~\cite{duda1972use} detects lines by aggregating the pixel intensity in the parameter space of lines and output detected straight lines via a voting procedure. Modern methods such as \cite{stephens1991probabilistic30,von2008lsd} detect lines based on local edge filtering. Recently, \cite{xue2019learning} utilizes deep neural networks to improve the performance of the conventional LSD line detection algorithm \cite{von2008lsd}. In our experiment, we will compare ours with this enhanced LSD algorithm \cite{xue2019learning}.  

\paragraph{Wireframe Parsing} The wireframe parsing task was first proposed in~\cite{huang2018learning}. The work provided a large-scale dataset with wireframe annotations, a baseline method, and a set of evaluation metrics. After that, \cite{zhou2019end} proposed an end-to-end solution and significantly improved the performance. \cite{zhang2019ppgnet} is a counterpart work of \cite{zhou2019end} which also introduces a new annotated dataset. Meanwhile, \cite{xue2020holistically} was a follow-up work of \cite{xue2019learning} which holds the state-of-the-art result. \cite{lin2020deep} designed a hough-transform based convolutional operator for the line detection task. In order to handle the topology of junctions and lines, a graph neural network based method \cite{meng2020lgnn} was proposed to tackle the wireframe task.  Recently, LETR \cite{xu2021line} a transformer based approach, was proposed for line segment without heuristics-driven intermediate stages for edge and junction proposal generation.
Furthermore, \cite{zhou2019learning} proposed a pipeline for reconstructing 3D wireframes from  2D images. Strictly speaking, line detection is not wireframe parsing as it does not detect junctions of multiple line segments. Nevertheless, we will use the same metric proposed for the wireframe to evaluate the quality of our line segments with the endpoints being viewed as the junctions. In particular, we will compare with the state-of-the-art method in this category \cite{xue2020holistically} in our experiments. 

\paragraph{Object Detection} Recently, the performance of line detection and wireframe parsing has all been boosted by an improvement in methods for object detection. Specifically,~\cite{zhou2019end} has inspired~\cite{ren2015faster} and many other two-stage detection methods such as~\cite{girshick2015fast}. After that, \cite{xue2020holistically} combined \cite{xue2019learning} with \cite{zhou2019end} has pushed the performance further. \cite{xue2020holistically} and \cite{zhou2019end} are two state-of-the-art methods from 2019 and 2020, respectively. Both of them drew inspiration from the object detection community and adopted a two stage strategy. Notice that the object detection community has  evolved from two-stage detector \cite{he2017mask,ren2015faster} to one-stage detector \cite{liu2016ssd, redmon2016you} or anchor free detector \cite{centernet, law2018cornernet}. Motivated by these work, in this paper, we propose a \ours{} network that detects line segments from images in one stage (see Figure.~\ref{fig:net}), which aims to achieve a better trade-off between speed and accuracy. 

During the preparation of this paper, we become aware of a very recent work~\cite{huang2020tp} that casts the line detection problem as a similar learning problem of predicting three parameters for each line segment. However, they have adopted a different parameterization and a rather different network design than ours. We will discuss the differences in the next section as well as compare their algorithm with ours in the experiments.

In addition, the incidence relation between wireframe parsing and object detection can be summarized as follows. If we consider the line and junction as object, wireframe detection becomes a special object detection. Even so, the wireframe detection still has two basic aspects of difference with object detection. First, the deformation and variance of the objects in image is much more severe than that of the wireframe in images. Secondly, the density of wireframe in image is much higher than that of the objects in images. Hence, less deformation or variance makes wireframe easy to recognize, but, higher density makes it easy to miss lines.

\section{Method}\label{sec:method}

The overall structure of the proposed network is illustrated in Figure~\ref{fig:net}. Given an input image, we first use a convolutional neural network to extract a shared feature map. Then the map is forwarded to separate sub-networks to predict three line representation maps: the line center map, the line length map, and the line angle map. These three maps are supervised by the pixel-wise loss between prediction and ground-truth. The network is optimized end-to-end with stochastic gradient descent. Below we give a detailed description and justification for each component. 

\subsection{Line Representation}\label{sec:method-linerep}

We represent a line segment by its center, length, and angle. Denote $\mathbf{p}_c \in \mathbb{R}^2$ as the center of line in the image coordinate, $\mathbf{p}_l$ and $\mathbf{p}_r \in \mathbb{R}^2$ as the left and right endpoints, respectively. We have the following relationship:
\begin{align}
  \mathbf{p}_l &= \mathbf{p}_c + \frac 1 2(\Ll \cos{\La}, \Ll \sin{\La}) \; \in \mathbb{R}^2, \\
  \mathbf{p}_r &= \mathbf{p}_c - \frac 1 2(\Ll \cos{\La}, \Ll \sin{\La}) \; \in \mathbb{R}^2, \label{eq::line}
\end{align}
where $\Ll$ is the length and $\La$ is the angle between line and the horizontal direction.

Any three of the above five quantities can uniquely determine a line segment. This leads to many mathematically equivalent representations of a line segment. For instance, the recent work \cite{huang2020tp} uses the center $\mathbf{p}_c$ and the $x$ and $y$ offsets of an end point to parameterize a segment. 

Among different choices, in this work, we choose to use the center $\mathbf{p}_c$, length $l$, and angle $\alpha$ for the following reasons: Firstly, the angle is the easiest one to predict since it can be identified accurately even from a local patch. It also has strong statistical priors (see Figure \ref{fig:anglehist}) that one can exploit to design more effective filters as we will elaborate more in the next subsection. In contrast to angle prediction, the network needs to perceive the whole line before making accurate predictions for length, center, and the endpoint (offsets). Secondly, since the relationship between line angle and the line's ending points is not one-to-one (one end point may be shared by multiple line segments), we choose to use line center and line length rather than the endpoints to simplify the inference. The resulting line representation gives several advantages:
\begin{enumerate}
    \item It naturally converts the line parsing problem to a pixel-wise classification/regression problem. Such transformation enables us to build a fully-convolutional structure for this task, which is both accurate and efficient (Section~\ref{sec::method}).
    \item Due to the pixel-wise formulation, it is not necessary to sample different kinds of lines as in~\cite{zhou2019end}, which significantly reduces the number of hyperparameters to tune (Section~\ref{sec::training}).
    \item The inference algorithm is straightforward. Given a predicted center location, we can directly use the corresponding predicted length and angle to get a line segment (Section~\ref{sec::inference}).
\end{enumerate}

\subsection{Fully Convolutional Line Parsing Networks}\label{sec::method}

\begin{figure}
\centering
\begin{minipage}[t]{0.33\linewidth}\centering
\includegraphics[width=0.79\linewidth]{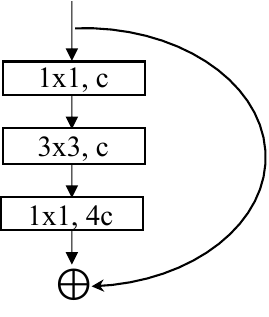} \\
{\small (a) conventional conv. block}
\end{minipage}
\begin{minipage}[t]{0.33\linewidth}\centering
\includegraphics[width=0.99\linewidth]{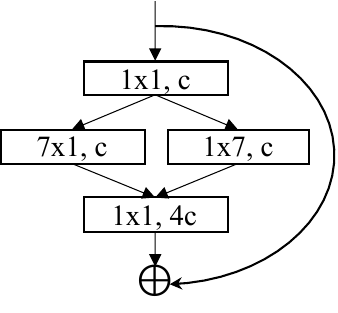} \\
{\small (b) customized conv. block}
\end{minipage}

\begin{minipage}[t]{0.33\linewidth}\centering
\includegraphics[width=0.95\linewidth]{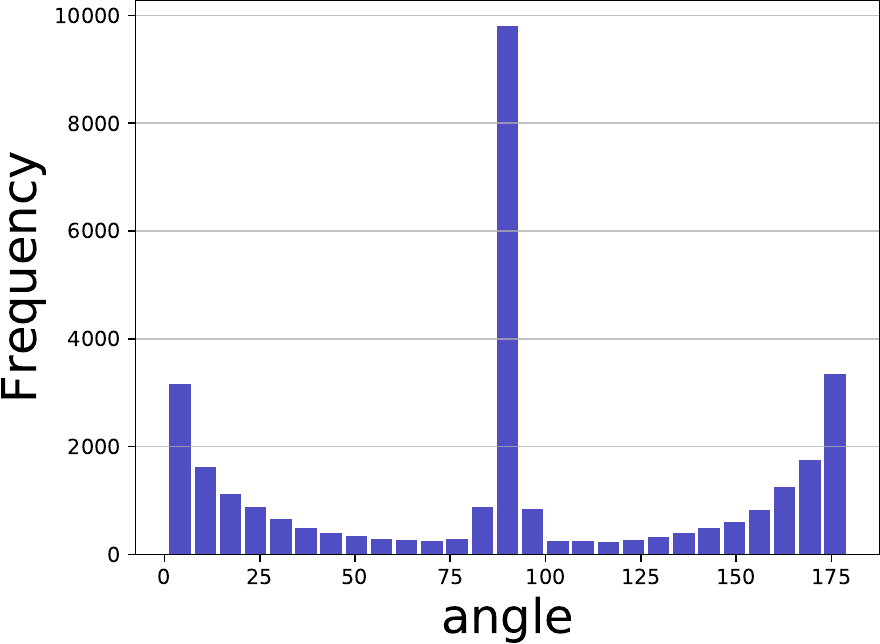} \\
{\small (c) angle histogram \cite{huang2018learning}}
\end{minipage}
\begin{minipage}[t]{0.33\linewidth}\centering
\includegraphics[width=0.95\linewidth]{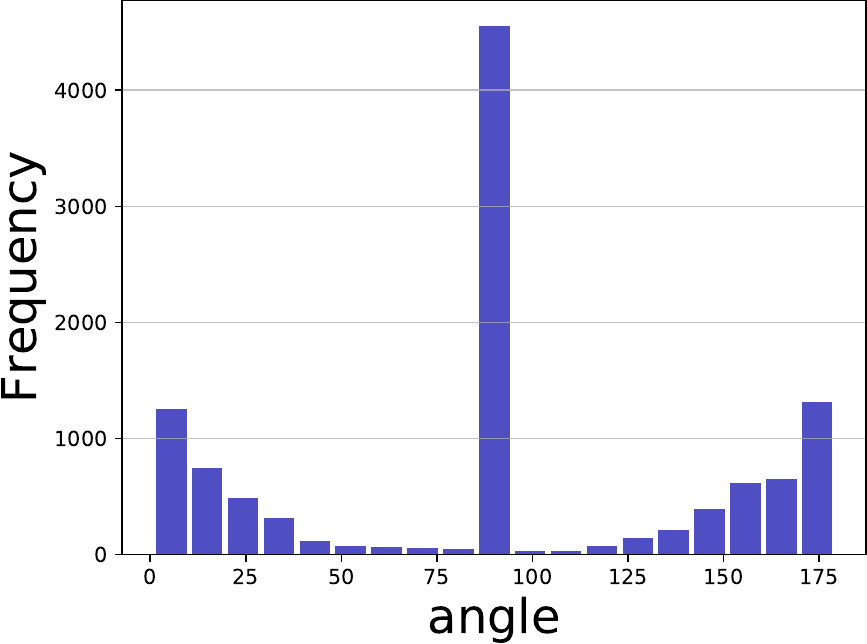} \\
{\small (d) angle histogram \cite{YorkUrban}}
\end{minipage}

\begin{minipage}[t]{0.33\linewidth}\centering
\includegraphics[width=0.95\linewidth]{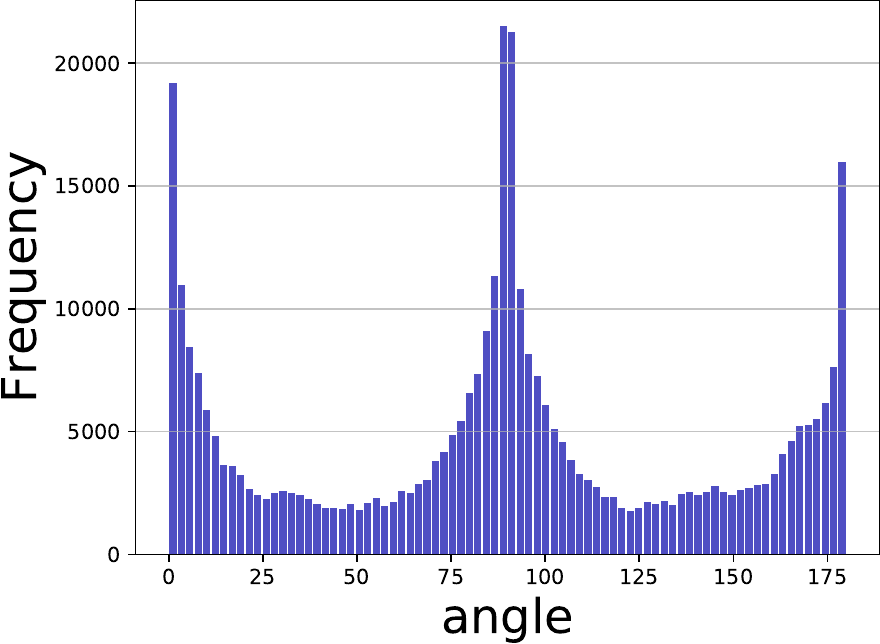} \\
{\small (e) angle histogram \cite{zhou2019learning}}
\end{minipage}

\caption{\small The top two are the convolutional blocks used in our method, conventional and customized, respectively. The (c), (d), (e) figures are the histograms of line angles in the ShanghaiTech dataset~\cite{huang2018learning}, the YorkUrban dataset~\cite{YorkUrban}, and the SceneCity dataset~\cite{zhou2019learning}. 
}
\label{fig:anglehist}
\end{figure}

\noindent\textbf{Designs of Backbone Networks.}
There are many advantages for our one-stage network. First, a one-stage network usually can improve the efficiency. Second, the simpler architecture allows easy customization. Line detection is often used in applications with very different requirements in speed and accuracy: e.g. the real-time localization or mapping versus offline 3D modeling from images. Some applications process one image at a time whereas others may process images in batches. Hence, in this work, we provide customized versions of our \ours{} along two lines of improvement. Along the first line, in order to speed-up the network, we simplify the backbone without sacrificing much performance. Along the second line, we show how to push the overall accuracy without compromising much speed (at least in the batch processing mode). 

First of all, following the work of \cite{newell2016stacked,xue2020holistically,zhou2019end}, we adopt an hourglass network with two stack modules as our default version \ours{} (HG2). We simplify the hourglass with one stack module to get a fast version \ours{} (HG1), then reduce the number of hierarchical structure in the hourglass block, from 4 to 3 and 2, to obtain two even faster versions \ours{} (HG1-D3) and \ours{} (HG1-D2).

The second line of improvement is to increase the accuracy without sacrifice too much speed. We observe from the datasets that most line segments in man-made environments are close to being vertical or horizontal (see Figure~\ref{fig:anglehist} (c), (d), (e)). Actually, this is typically the case for most real-world line detection tasks. The reason can be summarized as two aspects. First, in most man-made environments, the principle of parallel and vertical design is often used. Second, most people take pictures in a straight pose, maybe with a little ``pitch'' angle, but no ``roll'' angle. The above two reasons make the number of 0 and 90 degree lines dominant in all lines. Based on this strong statistical prior of the angle distribution, we customize the design of the line detection blocks (see Figure~\ref{fig:anglehist} (b)) to exploit such priors with similar computational cost (compared to that in Figure~\ref{fig:anglehist} (a)). This results in a more accurate version of \ours{} (HG2-LB) (where LB is short for line block). Moreover, to further improve the performance, we exploit multiple resolutions of the input through a parallel structure such as that in the high resolution (HR) network~\cite{sun2019deep}, which exhibits state-of-the-art performance for many vision tasks such image segmentation, detection, and recognition. This leads to a high-performance version of \ours{} (HR).

So overall, we have 6 different versions of \ours: 1) \ours{} (HG1-D2); 2) \ours{} (HG1-D3); 3) \ours{} (HG1); 4) \ours{} (HG2); 5) \ours{} (HG2-LB); and 6) \ours{} (HR). Their relative accuracy and speed are illustrated in Figure \ref{fig:teaser}, and their quantitative and qualitative evaluations will be given in Table \ref{tab:comparsion} and Figure \ref{fig:viz} in the experiment section.

\noindent\textbf{Line Score Maps.} We design the size of network output to be the same as the feature map size (128$\times$128) to make inference efficient. However, such a design will inevitably introduce quantization errors. We address this issue by introducing a local offset parameter associated with each line center. Specifically, for a ground-truth line center $\mathbf{p}_c$ in the original image, the line center score map \textit{C} is $1$ only in the $\lfloor \mathbf{p}_c / s \rfloor$ coordinates. For each center location, we also predict an offset value $\mathbf{p}_c / s - \lfloor \mathbf{p}_c / s \rfloor$, a line length value $l$, and a line angle value $\alpha$.

\noindent\textbf{Prediction Head.} In contrast to~\cite{huang2020tp} where different network structures are used for different tasks, we intentionally minimize the design effort of each prediction head. For each branch, given the shared feature map of channel dimension $c$, we use two 3$\times$3 convolutions with 256 output channels to refine the feature. Then we use a 1$\times$1 convolution with task-specific output channel(s) to predict each quantity. The number of output channels of heatmaps is 2, 2, 1, 1 for center, offset, length, and angle prediction, respectively.

\subsection{Training}\label{sec::training}

\noindent\textbf{Loss Function.}
We treat the problem of prediction \textit{C} as a classification problem and use the focal loss~\cite{lin2017focal} to tackle the unbalanced positive/negative data samples:
\begin{equation}
\mathcal{L}_{\textit{C}}\doteq -\frac{1}{N}\sum_{i,j}
\begin{cases}
C_{i,j}(1-\hat{C}_{i,j})^{\beta}\log(\hat{C}_{i,j}) & \text{if } C_{i,j}=1,\\
(1-C_{i,j})\hat{C}_{i,j}^{\beta}\log(1-\hat{C}_{i,j})  & \text{otherwise,}
\end{cases}
\end{equation}
where $\beta$ is the hyper-parameter of the focal loss, \textit{N} is the number of pixels in the score map, and $\hat{C}$ is the probability of each bin after softmax operation. 

Following~\cite{zhou2019end}, we use $\ell_{2}$ regression to predict the offset map $\textbf{O}$. The loss on $\textbf{O}$ is averaged over the the number of line center points in the score map:
\begin{equation}
\mathcal{L}_{\mathbf{O}}\doteq -\frac{1}{N}\sum_{i,j}
\| \hat{\mathbf{O}}_{i,j} - \mathbf{O}_{i,j} \|_2^2, 
\end{equation}

For line length and line angle prediction, the score map is normalized by sigmoid activation. Then we use the $\ell_1$ loss between the predicted and ground-truth value (as we empirically found that $\ell_1$ loss is better than $\ell_2$): 
\begin{equation}
\mathcal{L}_{\textit{L}}\doteq -\frac{1}{N}\sum_{i,j}
| \hat{L}_{i,j} - L_{i,j} |,\; 
\mathcal{L}_{\alpha}\doteq -\frac{1}{N}\sum_{i,j}
| \hat{A}_{i,j} - A_{i,j} |.
\end{equation}
The final loss used to train the network is:
\begin{align}\label{total_loss}
\mathcal{L}=\lambda_{\textit{C}}\mathcal{L}_{\textit{C}}+\lambda_{\textit{O}}\mathcal{L}_{\textit{O}}+\lambda_{\Ll}\mathcal{L}_{\Ll}+\lambda_{\La}\mathcal{L}_{\La}.
\end{align}

where $\lambda_{\textit{C}}$, $\lambda_{\textit{O}}$, $\lambda_{\Ll}$, $\lambda_{\La}$ mean the weights of each terms. We do the grid-search on the value of them and the best combination is $\lambda_{\textit{C}}=1$, $\lambda_{\textit{O}}=0.25$, $\lambda_{\Ll}=3$, $\lambda_{\La}=1$. Details can be found in section~\ref{sec::ablation-study}.

\noindent\textbf{Data Augmentation.} To make the model more robust to various viewpoints and scale, we perform the following data augmentations. In the first step, an image is processed with one of the following operations with equal probability:
\begin{enumerate}
    \item keep the original input image;
    \item flip it horizontally or vertically or simultaneously;
    \item rotate it by 90$^{\circ}$ clockwise or counterclockwise.
\end{enumerate}
After that, we use the random expansion augmentation in~\cite{liu2016ssd}. Specifically, we choose a $k\times k$ region in the 512$\times$512 zero input, and resize the image to fit inside the $k \times k$ region. We randomly sample $k$ from $[256, 512]$. This augmentation is to enhance  detection accuracy for short lines.

\subsection{Inference}\label{sec::inference}

During inference, we firstly apply a non-maximum suppression (NMS) on the line center score map to remove duplicate line detection results~\cite{zhou2019end}. Different from~\cite{zhou2019end}, we leverage the SoftNMS~\cite{bodla2017soft} from object detection to enhance the performance. Specifically, 
\begin{equation}\label{eq:SoftNMS}
  \textit{C'}_{i,j}  = \begin{cases}
    \textit{C}_{i,j} & \text{if }\textit{C}_{i,j} = \max_{(i',j')\in\mathcal{N}(i,j)} \textit{C}_{i',j'} \\
    \delta \cdot C_{i,j} & \text{otherwise}, \\
  \end{cases}
\end{equation}
where $\mathcal{N}(i,j)$ represents the 8 nearby bins around the location $(i, j)$. Such non-maximum suppression can be implemented with a max-pooling operator. After using SoftNMS, we use the top $K$ line centers according to their classification score. We use the corresponding predicted length and angle values to form a line according to Equation~\ref{eq::line}. 

The previous step only performs NMS on the point-level, without considering the effect of length and angle of a line. We hence propose a new structural NMS (StructNMS) that removes duplicate lines with the whole line structure. Starting from the line with the highest line center score (assume its index is $i$), we calculate the $\ell_2$ distance between its two endpoints and those of another line $j$:
\begin{equation}
\begin{aligned}\label{struc_nms}
d = \min \Big(&\| \mathbf{p}_l^i - \mathbf{p}_l^j \|_2^2 +  \| \mathbf{p}_r^i - \mathbf{p}_r^j \|_2^2, \\ 
&\| \mathbf{p}_l^i - \mathbf{p}_r^j \|_2^2 +  \| \mathbf{p}_r^i - \mathbf{p}_l^j \|_2^2\Big).
\end{aligned}
\end{equation}
Then we remove all the lines with $d$ less than a predefined threshold $\tau$. The procedure is applied for all the remaining line candidates.

\section{Experiments}\label{sec:exps}

In this section, we present our experiment results to analyze the performance of \ours{}, as well as compare with many other state-of-the-art line detection or wireframe parsing methods.

\subsection{Implementation Details}\label{sec::implementation}
\noindent\textbf{Details for Backbone Networks.} In order to make our neural network adaptable to various time efficiency requirement, \ours{} uses two different frameworks for the backbone networks: the stacked hourglass network \cite{newell2016stacked} for efficiency, and the HRNet \cite{sun2019deep} for performance. The stacked houglass network is a simple and elegant U-shape network that was previously used for wireframe parsing, such as in L-CNN~\cite{zhou2019end} and HAWP~\cite{xue2020holistically}. The configuration of our stacked houglass backbones are similar to the one in \cite{zhou2019end}. The main difference is that we provide five different settings to meet different efficiency needs: 1) one stack of the hourglass network with 2 hierarchical structure in hourglass block (\ours{} (HG1-D2)); 2) one stack of the hourglass network with 3 hierarchical structure in hourglass block (\ours{} (HG1-D3)); 3) one stack of the hourglass network with 4 hierarchical structures in the hourglass block (\ours{} (HG1)); 4) two stacks of the hourglass network with 4 hierarchical structures in hourglass block, while the one in L-CNN~\cite{zhou2019end} only provide the models containing 2 stacks of the hourglass networks with 4 hierarchical structures in hourglass block (\ours{} (HG2)); 5) two stacks of the hourglass network with 4 hierarchical structures in the hourglass block which residual blocks are all replaced with line block (see Figure~\ref{fig:anglehist} (b), \ours{} (HG2-LB)).

To further push the performance of \ours{}, we also employ the recent HRNet \cite{sun2019deep} as our backbone feature extractor (\ours{} (HR)).
HRNet is originally designed for human pose estimation tasks.
HRNet uses a more complex architecture design.
It starts with a high-resolution subnetwork (performing convolution on high-resolution feature maps) in its initial stages,  and gradually  add  some  low-resolution  subnetworks.
HRNet is designed to preserve more high-resolution details.
We use the HRNet-W32 variant \cite{sun2019deep} and find it performs better in \ours{} in term of accuracy, but it is much slower than the 2-stack hourglass network when batch size is equal to one.

\noindent\textbf{Prediction Head.} The prediction head transforms the feature maps from the backbone network into the final representations.
We simply use  two $3\times3$ convolution layers followed by a $1\times1$ convolution to match the output dimension.  All the convolution layers are activated with ReLU non-linearity.  The channel sizes of the middle $3 \times 3$ convolution are 128.

\noindent\textbf{Training.} We set a different initial learning rate $4 \times 10^{-4}$ and $4 \times 10^{-3}$ for the stack hourglass network and the high resolution network, respectively. Meanwhile, the parameter $\beta$ in focal loss is also different for the two backbones ($\beta=5$ for the stack hourglass network and $\beta=4$ for the high resolution network). {We choose the weights of the four loss terms in Equation~\eqref{total_loss} to be $\lambda_{\textit{C},\textit{O},\Ll,\La}=\{1, 0.25, 3, 1\}$. }
We train our neural network for 300 epochs, in which we decay the learning rate 10 times at the 240th epoch and the 280th epoch. All the experiments are conducted on a single NVIDIA GTX 2080Ti GPU. We use the ADAM optimizer \cite{kingma2014adam}. The weight decay is set to be $1 \times 10^{-4}$. We use a batch size that maximizes the occupancy of available GPU memory.

\noindent\textbf{Inference.} {There are two hyper-parameters in the inference stage that need to be determined ($\delta$ in Equation~\eqref{eq:SoftNMS} and $\tau$ in Equation~\eqref{struc_nms}). For $\delta$, we experimented with 10 numbers in the range from 0 to 1 (with a step 0.1) and chose the best $\delta=0.8$. For $\tau$, we experimented with 6 values (1, 2, 4, 8, 16, 32) and chose $\tau=2$ eventually.} 

\subsection{Experimental Settings}
\noindent\textbf{Datasets.}
We train and test \ours{} on the ShanghaiTech wireframe dataset \cite{huang2018learning}, which contains 5,000 training images and 462 testing images of man-made scenes.
We also include York Urban dataset \cite{YorkUrban}, a small dataset containing 102 images, as the testing dataset to evaluate the generalizability of different methods.

\noindent\textbf{Baselines.} We compare \ours{} with six baseline approaches: LSD \cite{von2008lsd},  DWP \cite{huang2018learning}, AFM \cite{xue2019learning},  L-CNN \cite{zhou2019end}, HAWP \cite{xue2020holistically}, and TP-LSD \cite{huang2020tp}.  Five approaches are supervised deep learning-based methods. To our best knowledge, they represent the state-of-the-art in their respective category of methods. We use the pre-trained models provided by the authors of each paper for evaluation, which are also trained on the ShanghaiTech wireframe dataset.

\noindent\textbf{Metric.}
The structural average precision (sAP)~\cite{zhou2019end}, proposed for evaluating accuracy in wireframe detection, uses the sum of squared error between the predicted end-points and their ground truths as evaluation metric. The predicted line segment will be counted as a true positive detection when its sum of squared error is less than a threshold, such as $\epsilon$ = 5, 10, 15.  AP$^H$ was used in wireframe parsing \cite{huang2018learning}. Instead of directly using the vectorized representation of line segments, we use heatmaps generated by rasterizing line segments for both parsing results and the groundtruth.

\subsection{Ablation Study} \label{sec::ablation-study}
In this section, we verify the effectiveness of our proposed method through extensive experiments. All of the experiments are performed on the ShanghaiTech dataset~\cite{huang2018learning} and structural AP are reported.

In Table~\ref{tab:ablation_train}, we analyze the choices of different training designs. Firstly, we show that using focal loss can improve the performance by about 1 point for all metrics, by comparing Table~\ref{tab:ablation_train} row (a) and (c). This is because the line centers only occupy a small portion of the image, thus the ratio between positive and negative samples is very small. In this case, focal loss is effective on addressing this problem. Secondly, we show the effectiveness of our proposed rotation and expansion data augmentation. In Table~\ref{tab:ablation_train} (b) to (g), the adding of rotation and expansion augmentation leads to an improvement of about 1 point and 3 points, respectively. These results show that by augmenting the data with different geometric transformations, one can obtain a more effective line detector that generalizes better.

\begin{table}[t]
\centering
\small
\setlength{\tabcolsep}{2.5pt}
\renewcommand{\arraystretch}{1.25}
\begin{tabular}{c|c|c|ccc|ccc}
    & \multirow{2}{*}{Backbone} & \multirow{2}{*}{$\mathcal{L}_{C}$} & \multicolumn{3}{c|}{DataAug} & \multirow{2}{*}{sAP${}^{5}$} & \multirow{2}{*}{sAP${}^{10}$} & \multirow{2}{*}{sAP${}^{15}$} \\
    & & & Flip & Rotate & Expand & & & \\
    \hline
    \hline
    (a) & \multirow{7}{*}{Hourglass} & CE & \checkmark & & & 56.2 & 61.4 & 63.2 \\
    (b) & & FL &  &  &  & 55.1 & 60.3 & 62.4 \\
    (c) & & FL & \checkmark & &  & 57.3 & 62.3 & 64.4 \\
    (d) & & FL & & \checkmark &  & 56.6 & 61.5 & 62.7 \\
    (e) & & FL & &  & \checkmark & 58.6 & 63.4 & 65.5 \\
    (f) & & FL & \checkmark & \checkmark &  & 58.4 & 63.3 & 65.3 \\
    (g) & & FL & \checkmark & \checkmark & \checkmark & 61.2 & 65.8 & 67.8 \\
    \hline
    (h) & HRNet & FL & \checkmark & \checkmark & \checkmark & 63.5 & 67.4 & 69.2 \\
\end{tabular}
\caption{\small Ablation study for our training details. CE and FL stand for cross-entropy loss and focal loss, respectively. DataAug refers to the data augmentation we used: flip, rotation, and expansion. We apply SoftNMS on the above models.}
\label{tab:ablation_train}
\end{table}

Next, we show the influence of different inference strategies. The results are shown in Table~\ref{tab:ablation_inference}. In entry (a), our method achieves 61.5 sAP${}^{5}$ using the original hard NMS in~\cite{zhou2019end}. Next, we apply the SoftNMS in Equation~\eqref{eq:SoftNMS} and the result improves by 2 points to 63.5. This is because at this stage, only point information is utilized. Thus there may be different lines with close center locations that are mistakenly removed. Setting a lower confidence instead of completely deleting such lines maintains the potential to recover such mistakes. Next, we show that using the StructNMS can further improve the performance by 1 point since such mechanism takes the whole line into consideration. Combining these two new NMS mechanisms improves 3 points over the original pipeline.

\begin{table}[t]
    \centering
    \small
    \setlength{\tabcolsep}{2.5pt}
    \renewcommand{\arraystretch}{1.05}
    \begin{tabular}{c|cc|ccc}
     & SoftNMS & StructNMS & sAP${}^{5}$ & sAP${}^{10}$ & sAP${}^{15}$\\
    \hline
    \hline
    (a) & & & 61.5 & 65.6 & 67.1 \\
    (b) & \checkmark & ~ & 63.5 & 67.4 & 69.2 \\
    (c) & & \checkmark & 62.5 & 66.7 & 68.3 \\
    (d) & \checkmark & \checkmark & 64.3 & 68.3 & 70.1 \\
    \end{tabular}
    \caption{\small Ablation study for inference details. The SoftNMS and StructNMS columns are two operators mentioned in methods. The model is an HRNet with focal loss and all the data augmentation.}
    \label{tab:ablation_inference}
\end{table}

Finally, we ablation study the weights of each terms in loss function. Firstly, setting the weights to 1, 0.25, 1, 1 as default, the value 0.25 of $\lambda_{\textit{O}}$ inherited from LCNN. We fix the $\lambda_{\textit{C}}=1$ since the $\alpha$ in focal-loss can play the same role as $\lambda_{\textit{C}}=1$. For the $\lambda_{\Ll}$ and $\lambda_{\La}$, we fix the $\lambda_{\La}=1$ and ablation study the value of $\lambda_{\Ll}$. The results shown in Table~\ref{tab:ablation_length} that 3 is the best value for $\lambda_{\Ll}$. We also fix the $\lambda_{\Ll}=1$ and ablation study the value of $\lambda_{\La}$, it turns out that $\lambda_{\La}=1$ is best.

\begin{table}[t]
    \centering
    \setlength{\tabcolsep}{2.5pt}
    \renewcommand{\arraystretch}{1.05}
    \begin{tabular}{c|cccc}
     $\lambda_{\Ll}$ &  1   &  2   &  3   & 4 \\
    \hline
    \hline
    sAP${}^{5}$      & 61.8 & 63.5 & \textbf{64.3} & 63.9 \\
    \end{tabular}
    \caption{\small Ablation study on the weight of length $\lambda_{\Ll}$ where $\lambda_{\textit{C}}=1$, $\lambda_{\textit{O}}=0.25$, $\lambda_{\La}=1$.}
    \label{tab:ablation_length}
\end{table}

\begin{table*}
    \centering
    \footnotesize
    \setlength{\tabcolsep}{4.5pt}
    \renewcommand{\arraystretch}{1.25}
    \begin{tabular}{l|ccc|ccc|c|c}
    \multirow{2}{*}{Method} & \multicolumn{3}{c|}{ShanghaiTech} & \multicolumn{3}{c|}{YorkUrban} & FPS & FPS \\
    \cline{2-7}
     ~  & sAP${}^{5}$ & sAP${}^{10}$ & AP${}^{H}$ & sAP${}^{5}$ & sAP${}^{10}$ & AP${}^{H}$ & batch-size=1 & batch-size=10 \\
    \hline
    \hline
    \textit{Two-stage methods}      &  ~   & ~    & ~    & ~    & ~    & ~     & ~    & ~ \\
    LSD (320) \cite{von2008lsd}     & 6.7  & 8.8  & 52.0 & 7.5  & 9.2  & 51.0  & 100  & / \\
    AFM \cite{xue2019learning}      & 18.5 & 24.4 & 69.2 & 7.3  & 9.4  & 48.2  & 13.5 & / \\
    DWP \cite{huang2018learning}    & 3.7  & 5.1  & 67.8 & 1.5  & 2.1  & 51.0  & 2.24 & /  \\
    L-CNN \cite{zhou2019end}        & 58.9 & 62.9 & 80.3 & 24.3 & 26.4 & 58.5  & 15.6 & 16.4  \\
    HAWP \cite{xue2020holistically} & 62.5 & 66.5 & 84.5 & 26.1 & 28.5 & 60.6  & 26.9 & 81.3   \\
    HAWP (re-trained)               & 63.1 & 66.9 & 84.9 & 26.9 & 29.2 & 61.4  & 12.3 & 60.6   \\
    \hline
    \hline
    \textit{One-stage methods} & ~ & ~ & ~ & ~ & ~ & ~ & ~ & ~ \\
    TP-LSD (Res34-Lite)            & 56.4 & 59.7  & / & 24.8 & 26.8 & / & 65.9 & 327.6 \\
    TP-LSD (Res34) \cite{huang2020tp}       & 57.5 & 60.0  & / & 25.3 & 27.4 & / & 32.6 & 194.2 \\
    TP-LSD (re-trained)            & 62.4 & 67.3 & 85.2 & 27.5 & 39.8 & 63.1   & 17.1 & 106.2 \\
    \ours{} (HG1-D2)  & 52.7         & 57.2          & 76.7          & 23.9          & 26.1          & 56.3 & \textbf{73.4} & \textbf{363.1} \\
    \ours{} (HG1-D3)  & 57.8         & 62.7          & 82.0          & 25.6          & 28.2          & 60.6 & 67.9 & 339.8 \\
    \ours{} (HG1)     & 58.6         & 63.6          & 83.0          & 24.9          & 27.4          & 60.3          & 64.1 & 324.5  \\
    \ours{} (HG2)     & 61.3         & 65.8          & 84.0          & 27.2          & 29.4          & 62.0          & 35.7 & 205.2  \\
    \ours{} (HG2-LB)  & 62.6         & 66.8          & 85.1          & 27.6          & 29.9          & 62.3          & 29.8 & 148.5  \\
    \ours{} (HR)      &\textbf{64.3} & \textbf{68.3} & \textbf{85.7} & \textbf{28.5} & \textbf{30.8} & \textbf{65.0}    & 17.4 & 107.5  \\
    \end{tabular}      
    \vspace{-0.1in}
    \caption{\small Quantitative Results and Comparisons. The proposed \ours{} achieves state-of-the-art results consistently. Overall, the FPS of our \ours's~is still significantly better than or on par with that of the six existing methods. Note that for fair apple-to-apple comparison, we have retrained the HAWP model and TP-LSD model using their latest released code and the same settings used in our \ours{} (HR). See text for details.}
    \label{tab:comparsion}
\end{table*}

\subsection{Comparison with Other Methods}

To ensure fair comparison with the previous state-of-the-art method HAWP~\cite{xue2020holistically}, we have re-implemented their method with slightly better results. We have adopted the same hyper-parameter settings as our best performing model \ours{} (HR), including the focal loss, backbone, longer training epoch, and data augmentation etc. The experiment details can be found in the section 4.5. We also re-implement the other one-stage method TP-LSD~\cite{huang2020tp}, using the same setting as our best performing model \ours{} (HR), the results can be seen in Table~\ref{tab:comparsion} ``TP-LSD (retrained)''.

In order to make fair comparison on speed, we compute the frames per second (FPS) for different methods with their latest released code using a single GPU (RTX 2080Ti). Most of the previous methods focus on the \textit{latency} of the algorithm (FPS while batch-size=1) but not \textit{throughput} (FPS while batch-size=max) which is a more important metric for offline batch processing. We show the throughput metric in the last column of Table~\ref{tab:comparsion} which illustrates the better parallel performance of single stage methods compare with two stage methods.

Table~\ref{tab:comparsion} summarizes our results. Our \ours{} obtains state-of-the-art performance both in terms of \textit{efficiency} and \textit{accuracy}. Under the very challenging sAP${}^{5}$ metric, using the same backbone network, our \ours{} (HG2) achieves comparable performance with previous state-of-the-art methods while being 1.4x faster when batch size is equal to one. By changing the backbone from HG2 to HG2-LB, we can get another 1.5 points gain without sacrifice too much speed. Moreover, our\ours{} (HR) achieves state-of-the-art with a decent speed (17.4 FPS) and achieves 1.8 times higher throughput than HAWP~\cite{xue2020holistically}. 
Specifically, our \ours{} (HG1) is 1 points higher than another one-stage method TP-LSD, while is 1.8 times faster. The re-implement version TP-LSD (retrained) shows the big gain of our framework (around 5 points improvement compare with TP-LSD (Res34)). When we reduce the hierarchical structure in hourglass block to 2, the speed further increases to 73 FPS. Our method not only achieves comparable speed with LSD, but is more than 8 times accurate in terms of sAP${}^{5}$. Finally, \ours{} also achieves the state-of-the-art results on YorkUrban dataset which shows its generalizability.

\begin{figure*}
    \centering
    \begin{minipage}[t]{0.24\linewidth}\centering
    \includegraphics[width=0.99\linewidth]{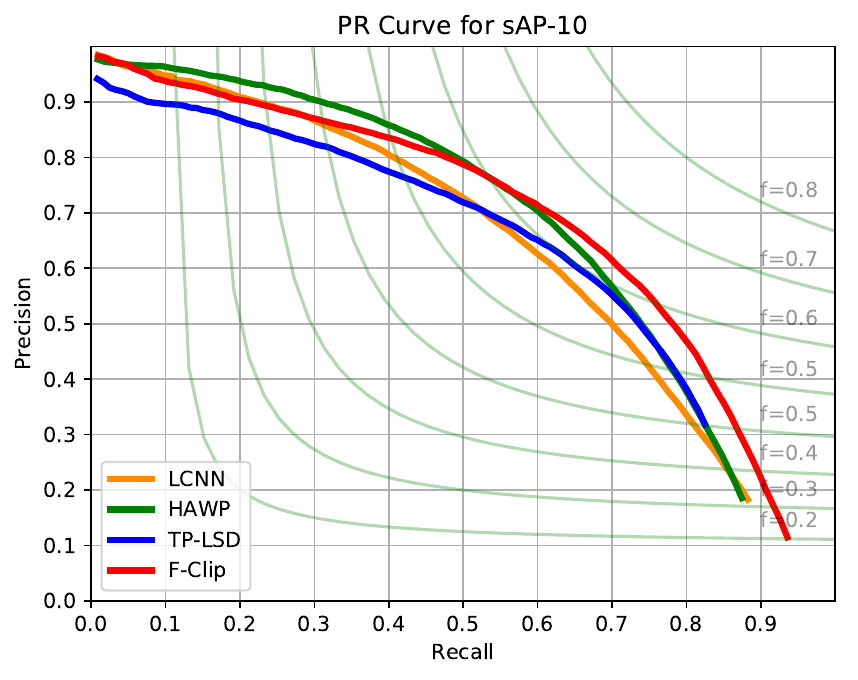}
    \small (a) sAP${}^{10}$ on ShanghaiTech
    \end{minipage}
    \begin{minipage}[t]{0.24\linewidth}\centering
    \includegraphics[width=0.99\linewidth]{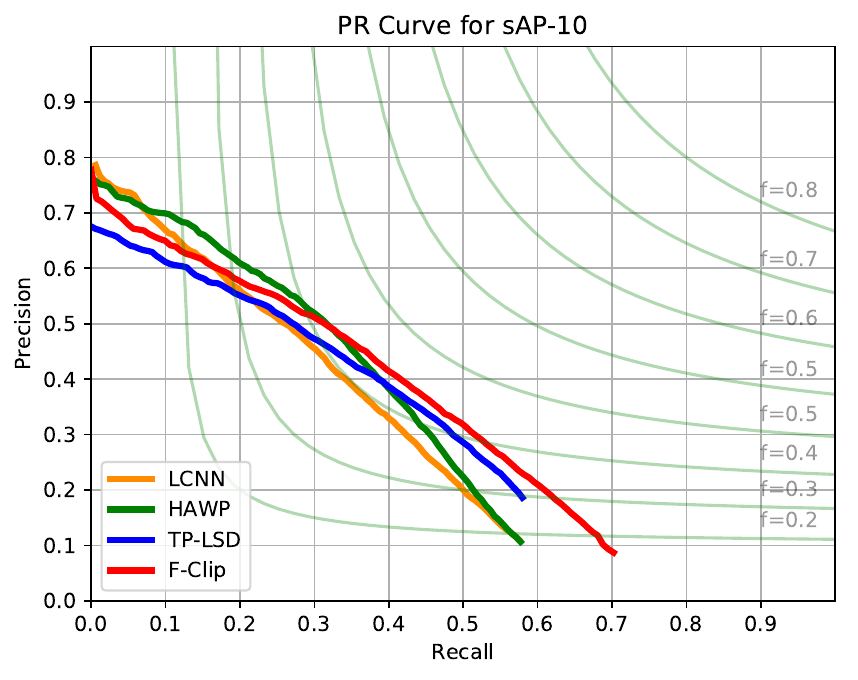}
    \small (b) sAP${}^{10}$ on YorkUrban
    \end{minipage}
    \begin{minipage}[t]{0.24\linewidth}\centering
    \includegraphics[width=0.99\linewidth]{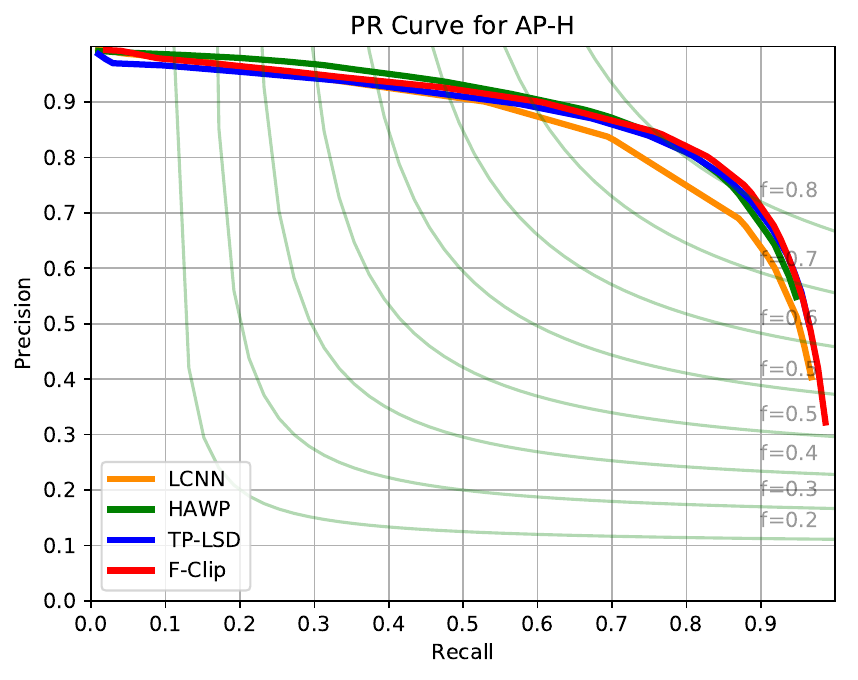}
    \small (c) AP${}^{H}$ on ShanghaiTech
    \end{minipage}
    \begin{minipage}[t]{0.24\linewidth}\centering
    \includegraphics[width=0.99\linewidth]{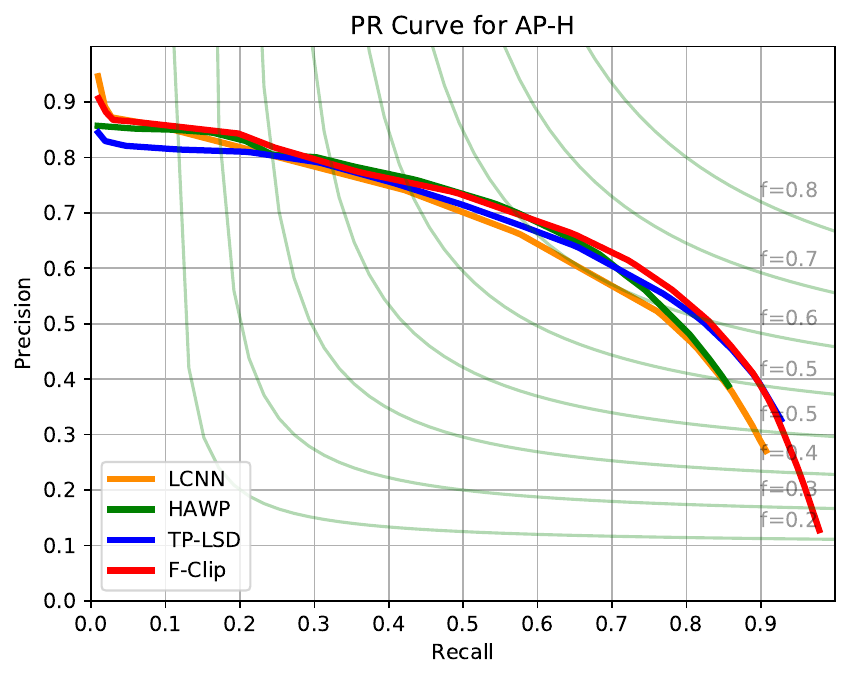}
    \small (d) AP${}^{H}$ on YorkUrban
    \end{minipage}
    \caption{\small Precision-Recall (PR) curves of sAP${}^{10}$ and AP${}^{H}$ for TP-LSD \cite{huang2020tp}, L-CNN \cite{zhou2019end}, HAWP \cite{xue2020holistically} and \ours{} (HR) on the ShanghaiTech and the YorkUrban benchmarks respectively. }
    \label{fig:pr-curve}
\end{figure*}

The precision and recall curves of sAP${}^{10}$ and AP${}^{H}$ are shown in Figure~\ref{fig:pr-curve}. For the ShanghaiTech dataset, our method achieves higher recall and performs better in the higher recall regime, similarly for the YorkUrban Dataset.

\subsection{Comparison with State-of-art Methods}
To make a fair comparison with the previous state-of-the-art method HAWP~\cite{xue2020holistically}, we adopt the hyper-parameter settings including the 1) backbone, 2) longer training epoch, 3) data augmentation, and 4) focal loss on HAWP same as our best performance model \ours~(HR).

\noindent\paragraph{Backbone.}
Our method employ a strong backbone network HRnet \cite{sun2019deep} (short for HR in Table~\ref{tab:2stage_long}). As shwon in Table~\ref{tab:2stage_long}, the HRnet do not bring significantly performance improvement for HAWP.

\noindent\paragraph{Training Epochs.}
Our method needs more training iterations to converge because we use a strong backbone network.  
As shown in Table~\ref{tab:2stage_long} below, additional training epochs do not improve significantly the performance of state-of-the-art two-stage methods HAWP. 

\noindent\paragraph{Data Augmentation.}
As shown in single-stage object detection methods~\cite{liu2016ssd,centernet}, applying a more complex data augmentation does not improve the performance of two-stage networks.  We also see the same phenomena, as shown in Table~\ref{tab:2stage_long}.

\noindent\paragraph{Focal Loss.} 
Focal loss~\cite{lin2017focal} is designed to handle the balance between positive and negative samples. We apply the focal loss on the junction detector of HAWP. As shown in Table~\ref{tab:2stage_long} below, the focal loss makes a bad effect on the performance of HAWP. 

\noindent\paragraph{Analysis.} 
Both two-stage wireframe detection methods LCNN \cite{zhou2019end} and HAWP are junction based methods. The performance of junction detection will dominate the performance of overall wireframe detection. Our \ours~is a single stage method which skip the detection of junction and predicts the line directly. Compare with line detection, the local feature is enough for the detection of junction. Hence, hourglass backbone with a short training epoch (30 epochs) is enough for converging to a good result, a strong backbone HRnet with a longer training epoch (300 epoch) does not bring significant performance improvement for HAWP. Meanwhile, focal loss on the junction detector even makes a bad effect on the performance of HAWP.

\begin{table}[!h]
    \centering
    \small
    \setlength{\tabcolsep}{3.5pt}
    \renewcommand{\arraystretch}{1.2}
    \begin{tabular}{c|c|c|c|c|c}
    Method & backbone & epoch & DataAug & focal-loss & sAP$^{5}$ \\
    \hline
    \hline
    \multirow{5}{*}{HAWP} & HG & 30  & ~          & ~          & 62.5 \\
     ~                    & HG & 300 & ~          & ~          & 62.8 \\
     ~                    & HR & 300 & ~          & ~          & 63.1 \\
     ~                    & HR & 300 & \checkmark & ~          & 63.0 \\
     ~                    & HR & 300 & \checkmark & \checkmark & 62.4 \\                    
     \hline
    Ours                  & HR & 300 & \checkmark & \checkmark & 64.5 
    \end{tabular}
    \vspace{8pt}
    \caption{HAWP with longer training epochs, hrnet backbone, focal loss, and the same data augmentation as ours.}
    \label{tab:2stage_long}
\end{table}

\subsection{Visualization}

\begin{figure*}
    \begin{minipage}[t]{0.19\linewidth}\centering
    \includegraphics[width=0.99\linewidth]{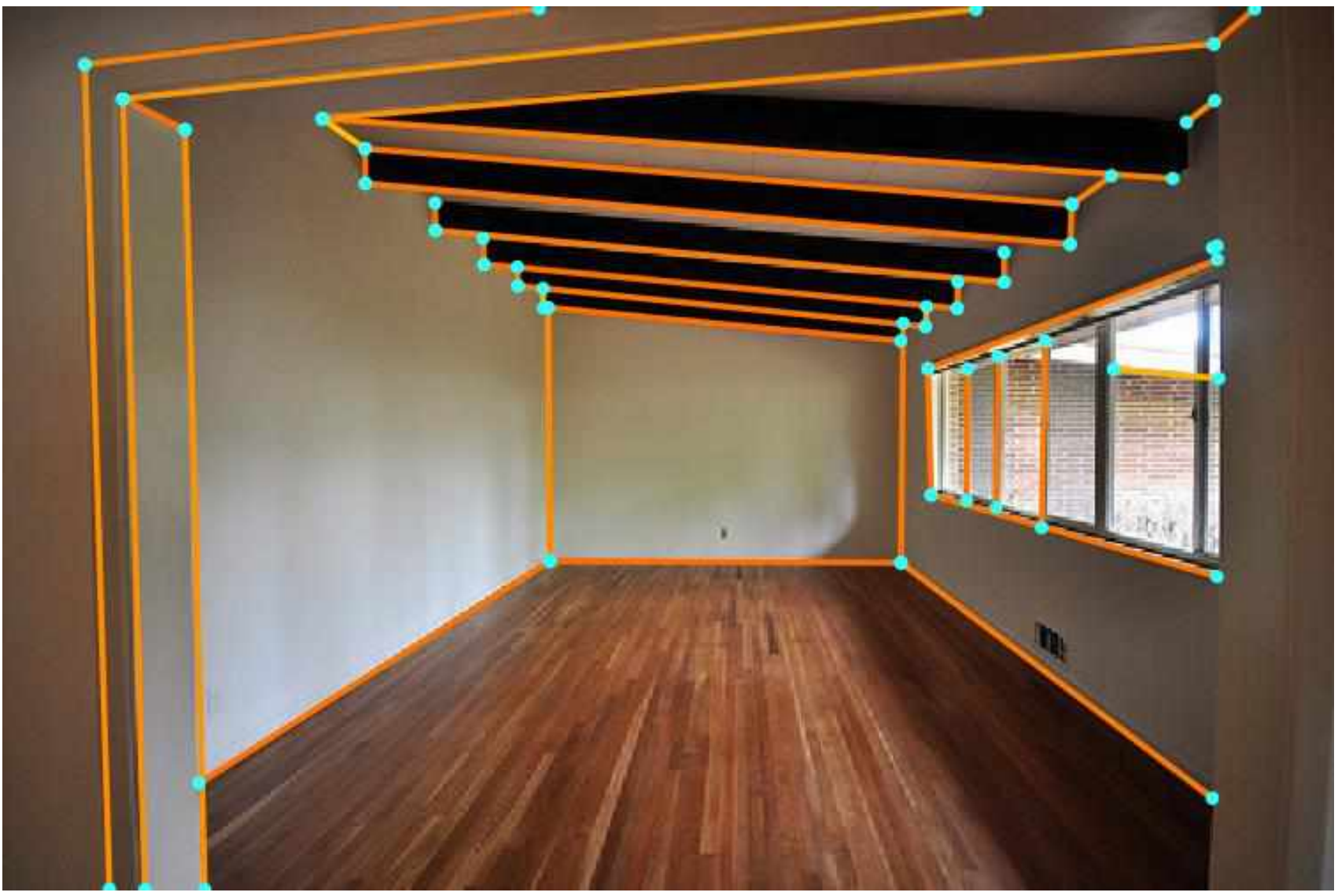}
    \includegraphics[width=0.99\linewidth]{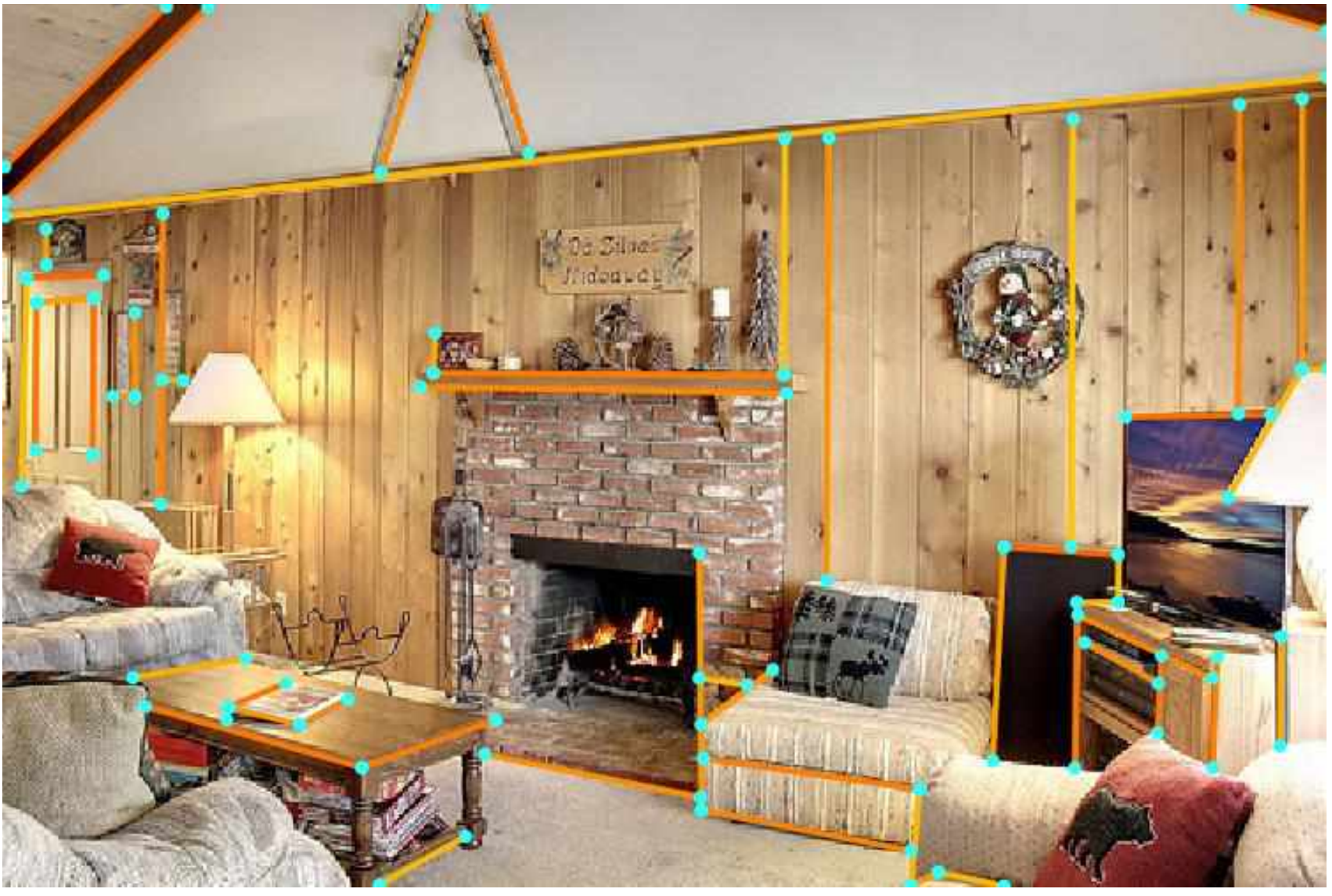}
    \includegraphics[width=0.99\linewidth]{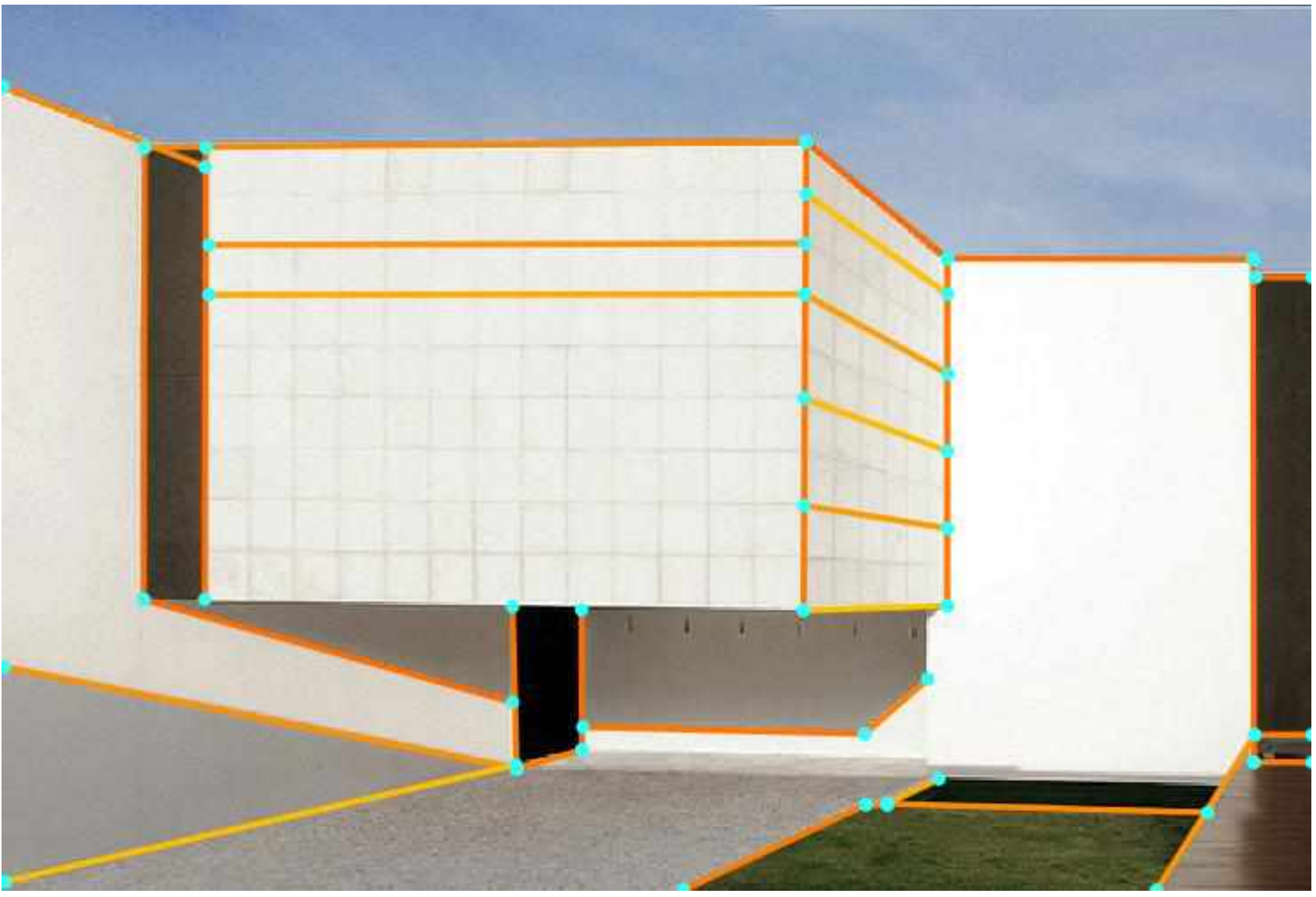}
    \includegraphics[width=0.99\linewidth]{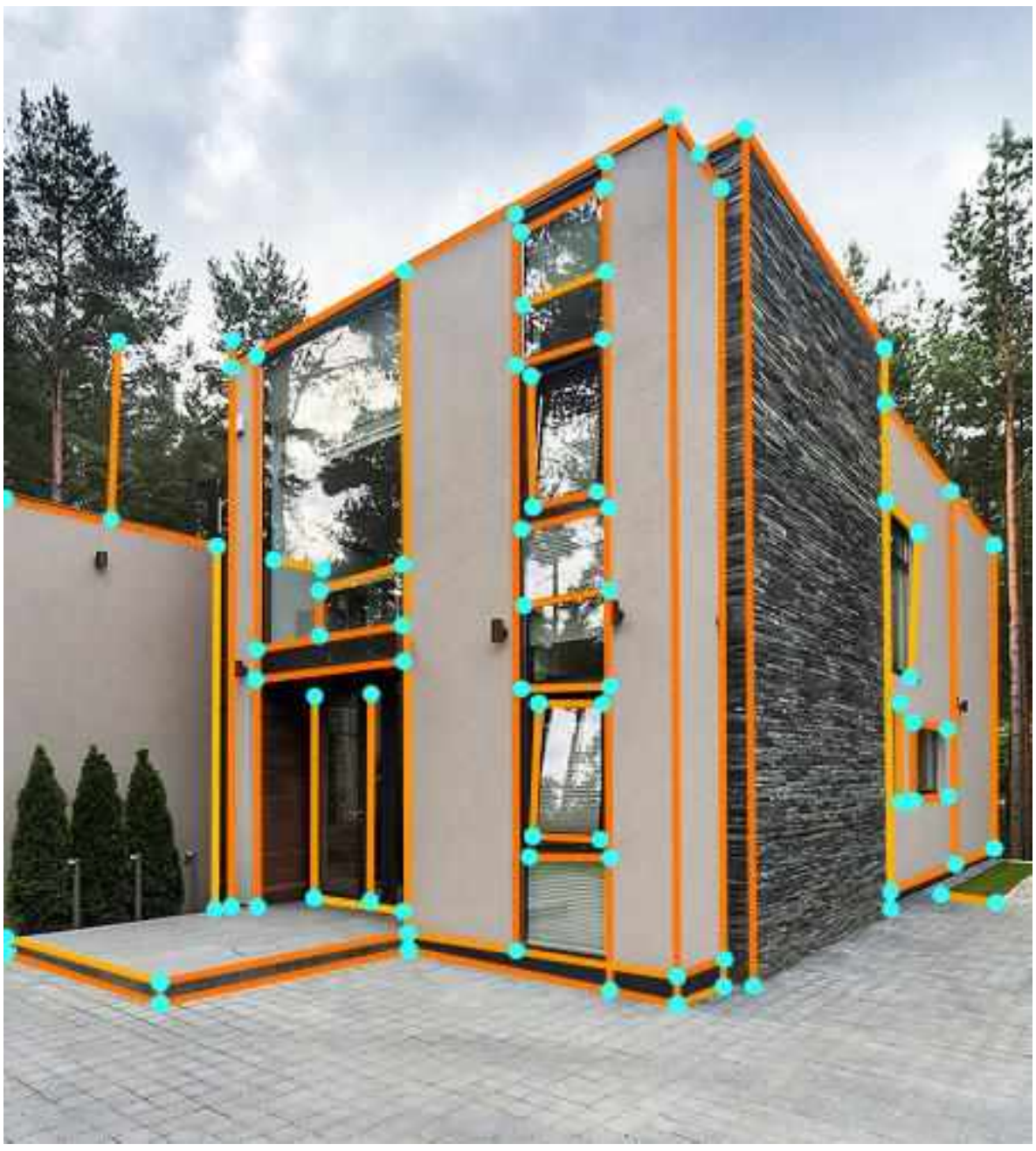}
    \includegraphics[width=0.99\linewidth]{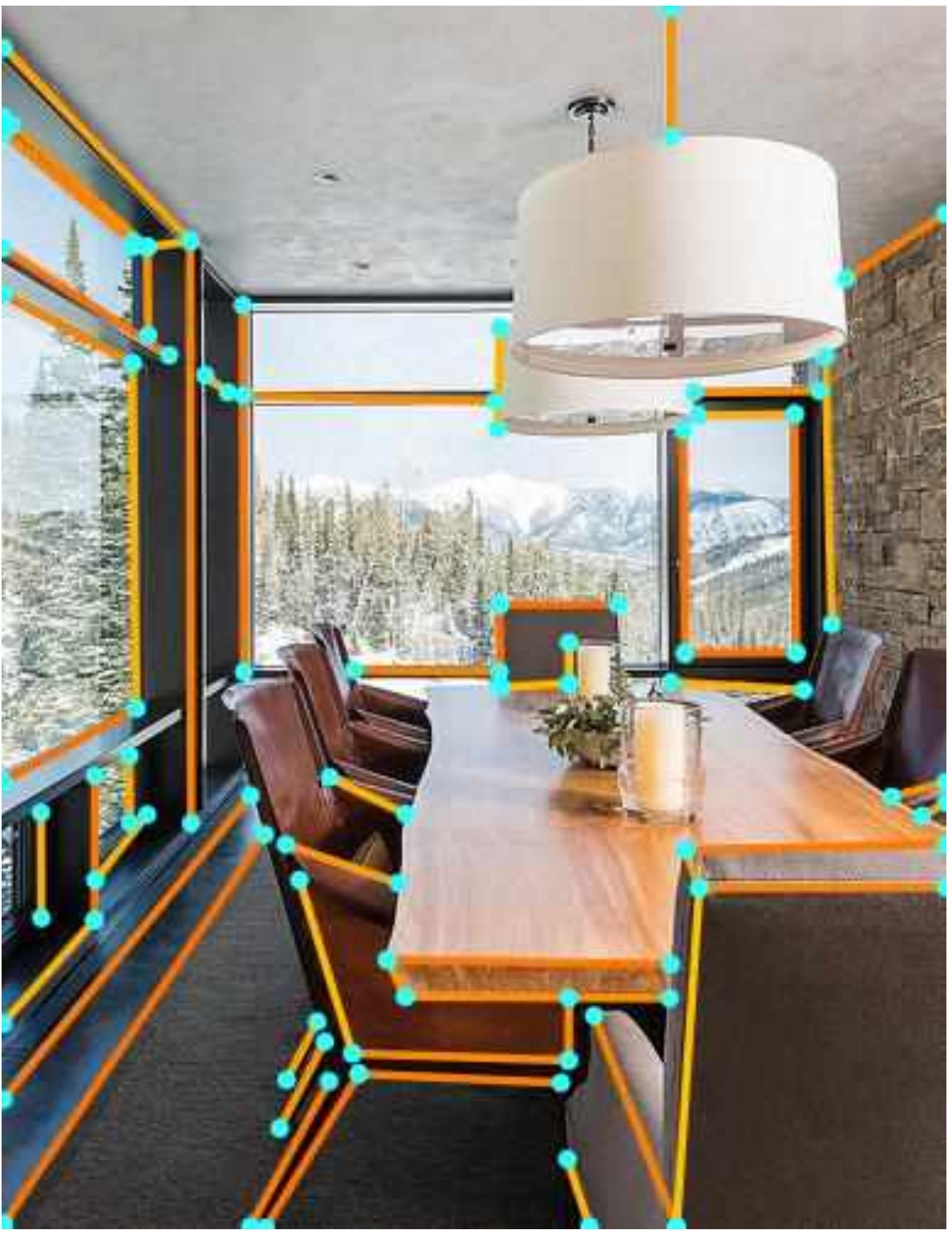}
    \includegraphics[width=0.99\linewidth]{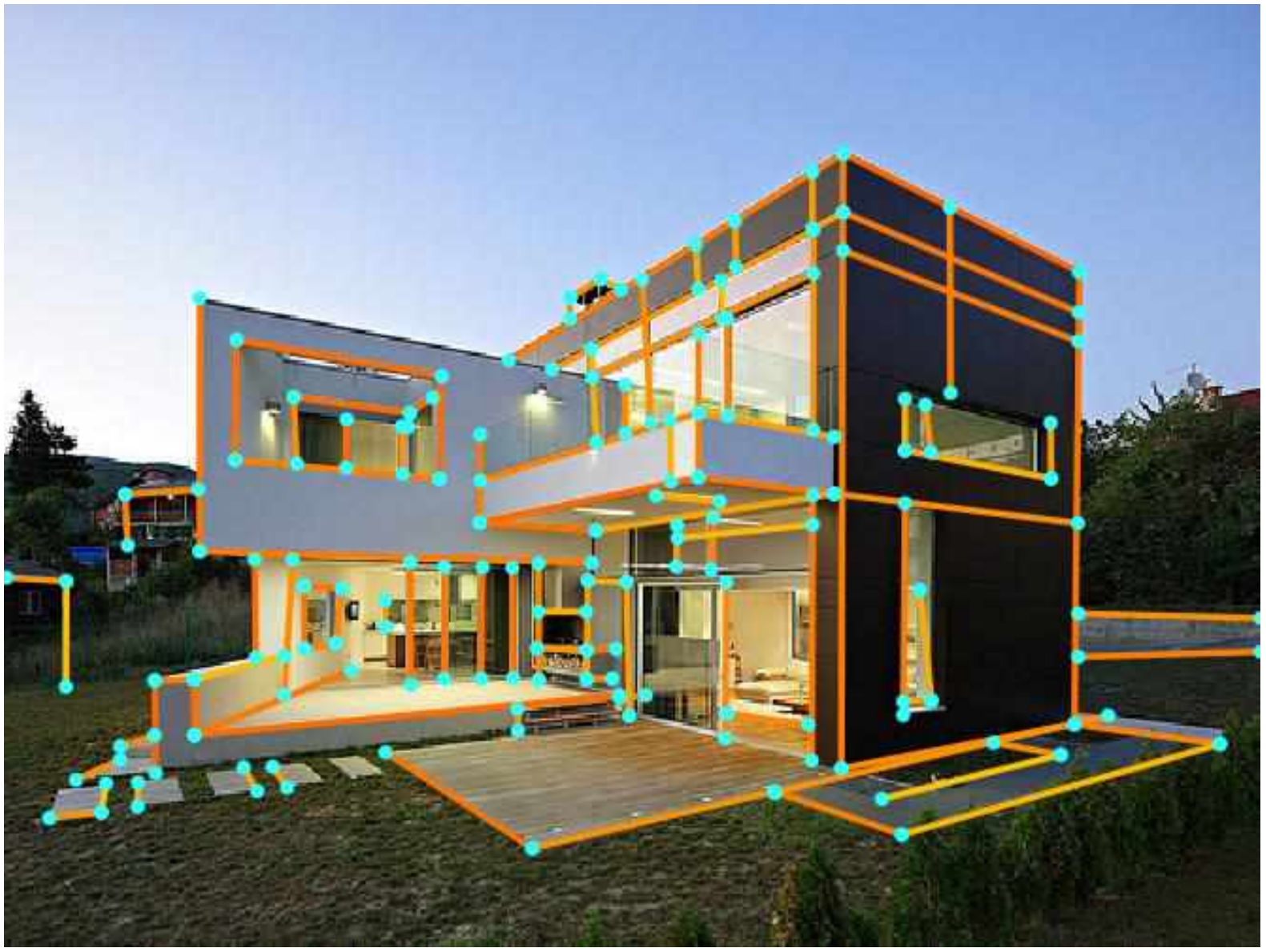}
    \includegraphics[width=0.99\linewidth]{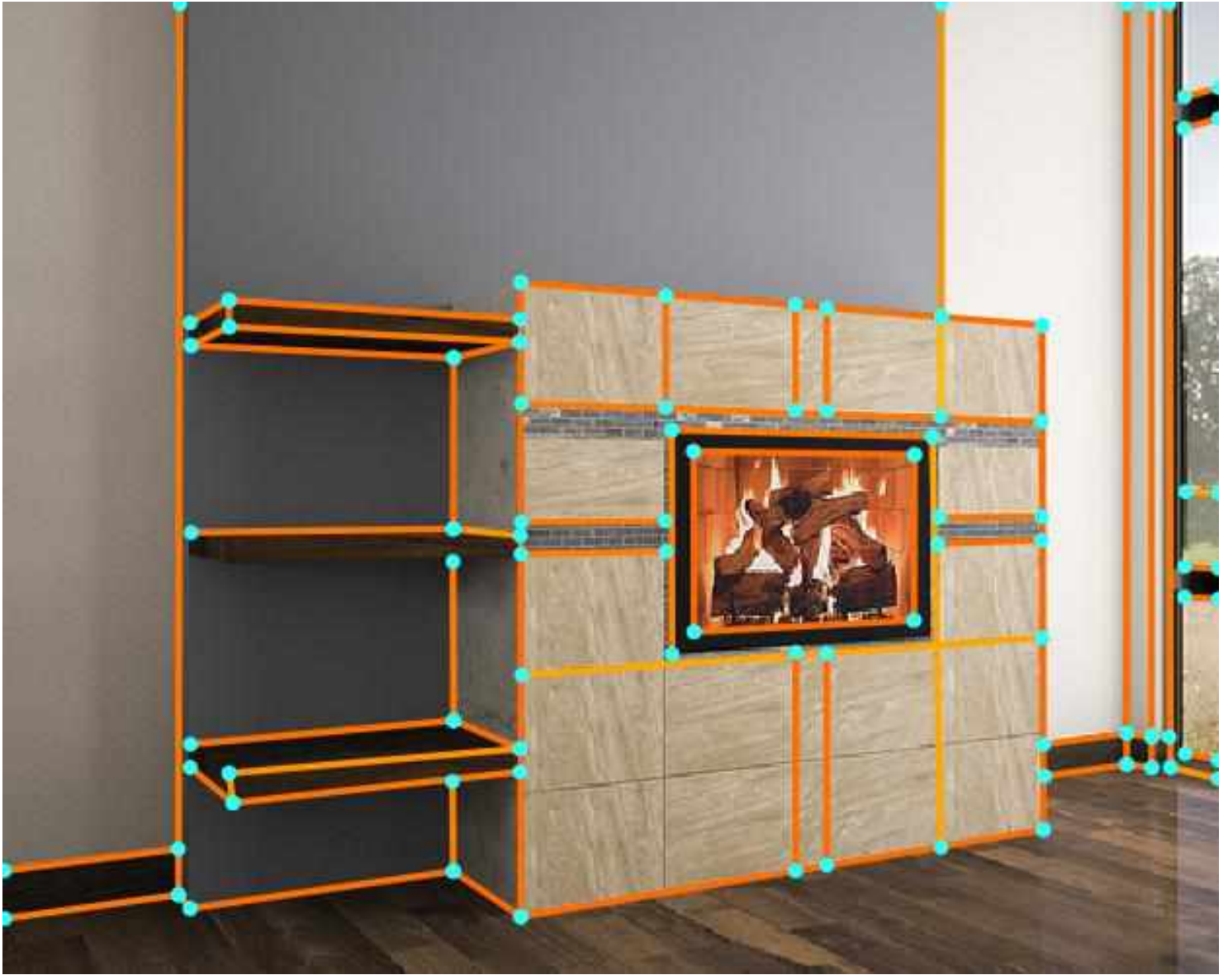}
    (a) L-CNN
    \end{minipage}
    \begin{minipage}[t]{0.19\linewidth}\centering
    \includegraphics[width=0.99\linewidth]{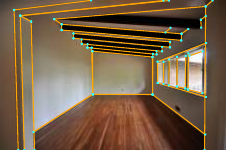}
    \includegraphics[width=0.99\linewidth]{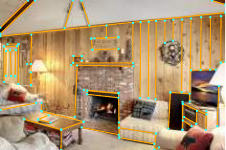}
    \includegraphics[width=0.99\linewidth]{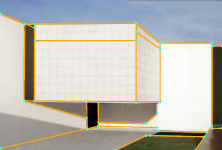}
    \includegraphics[width=0.99\linewidth]{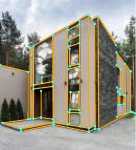}
    \includegraphics[width=0.99\linewidth]{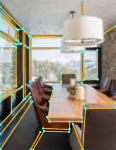}
    \includegraphics[width=0.99\linewidth]{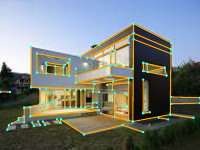}
    \includegraphics[width=0.99\linewidth]{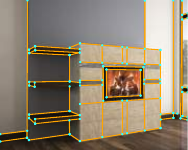}
    (b) HAWP
    \end{minipage}
    \begin{minipage}[t]{0.19\linewidth}\centering
    \includegraphics[width=0.99\linewidth]{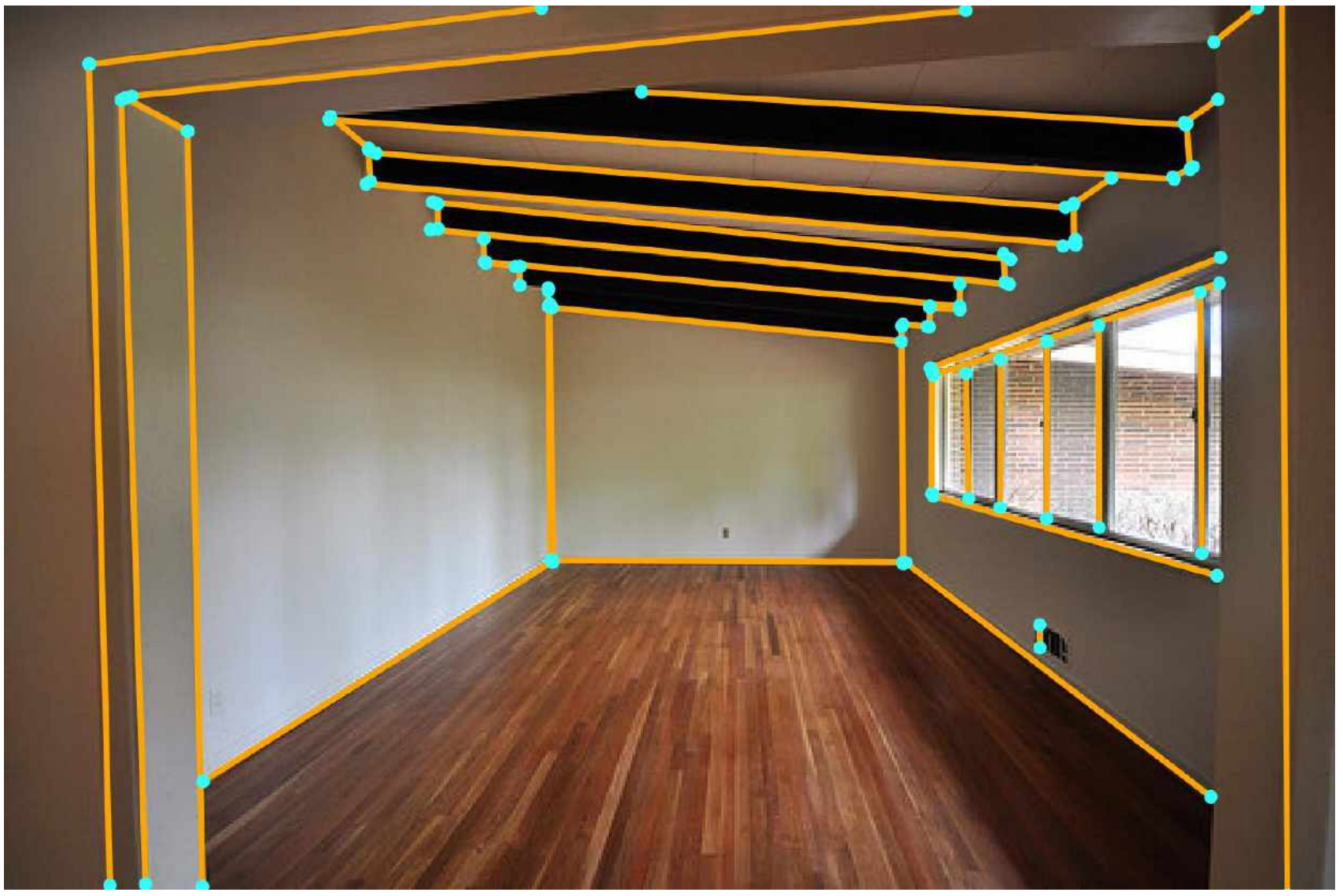}
    \includegraphics[width=0.99\linewidth]{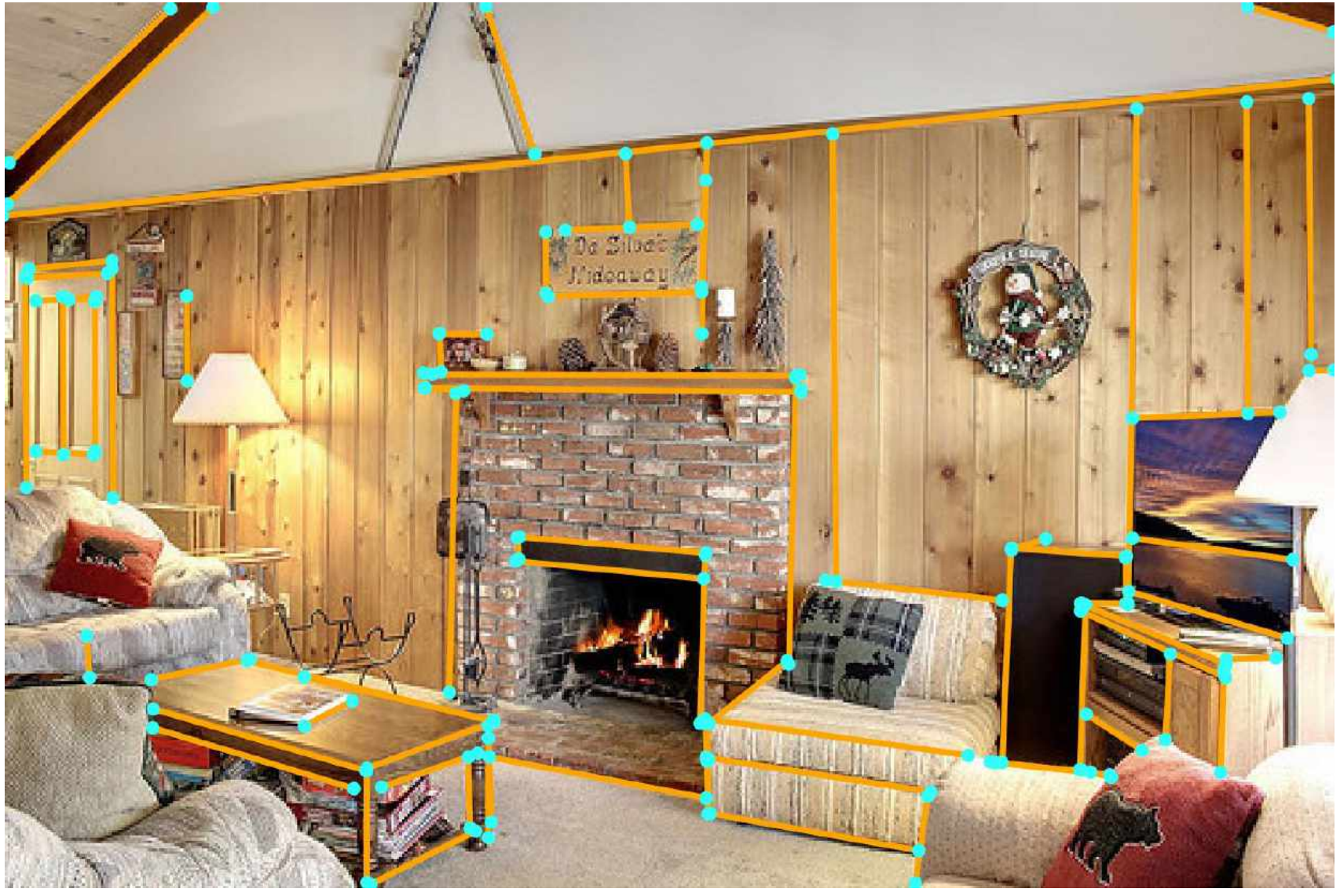}
    \includegraphics[width=0.99\linewidth]{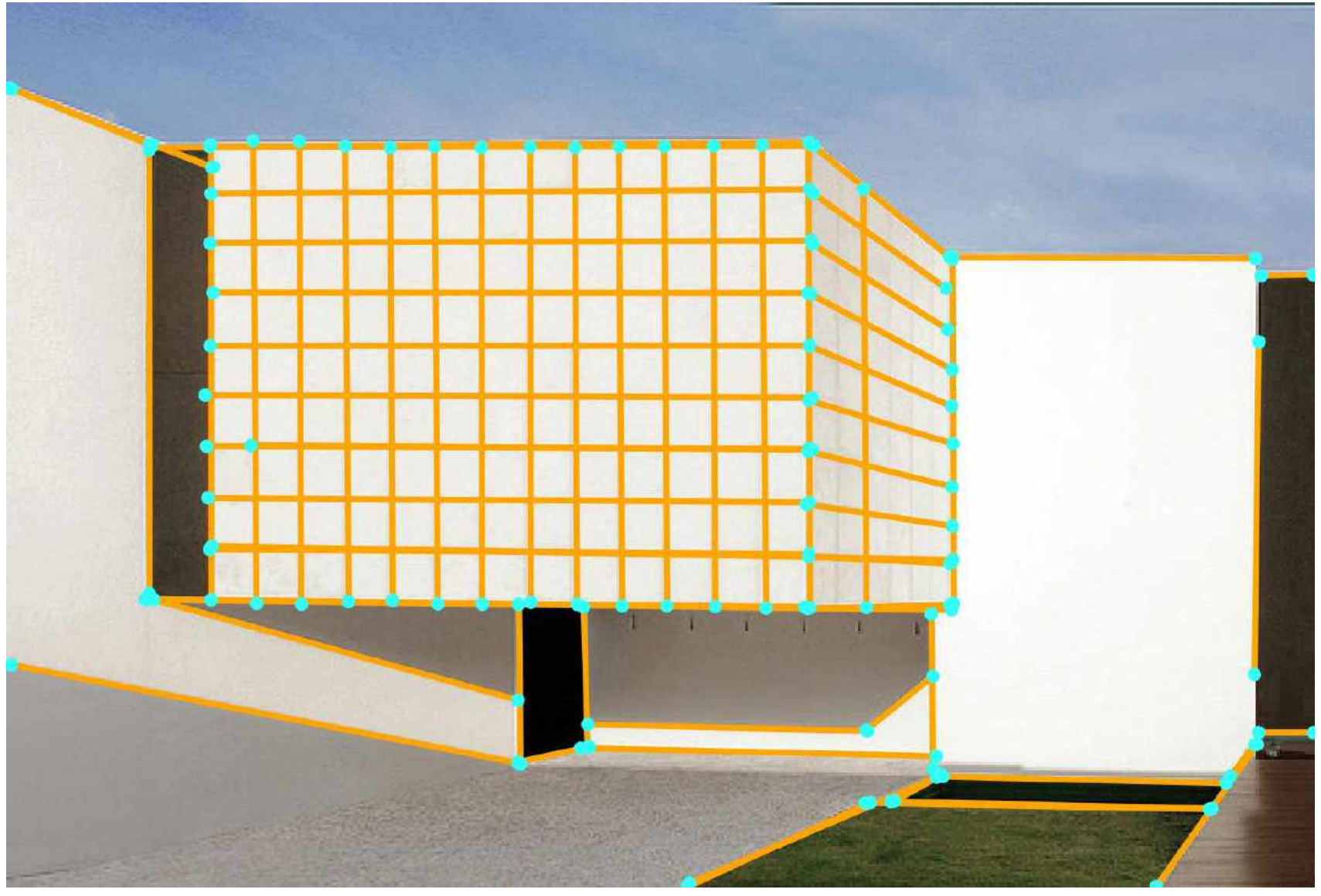}
    \includegraphics[width=0.99\linewidth]{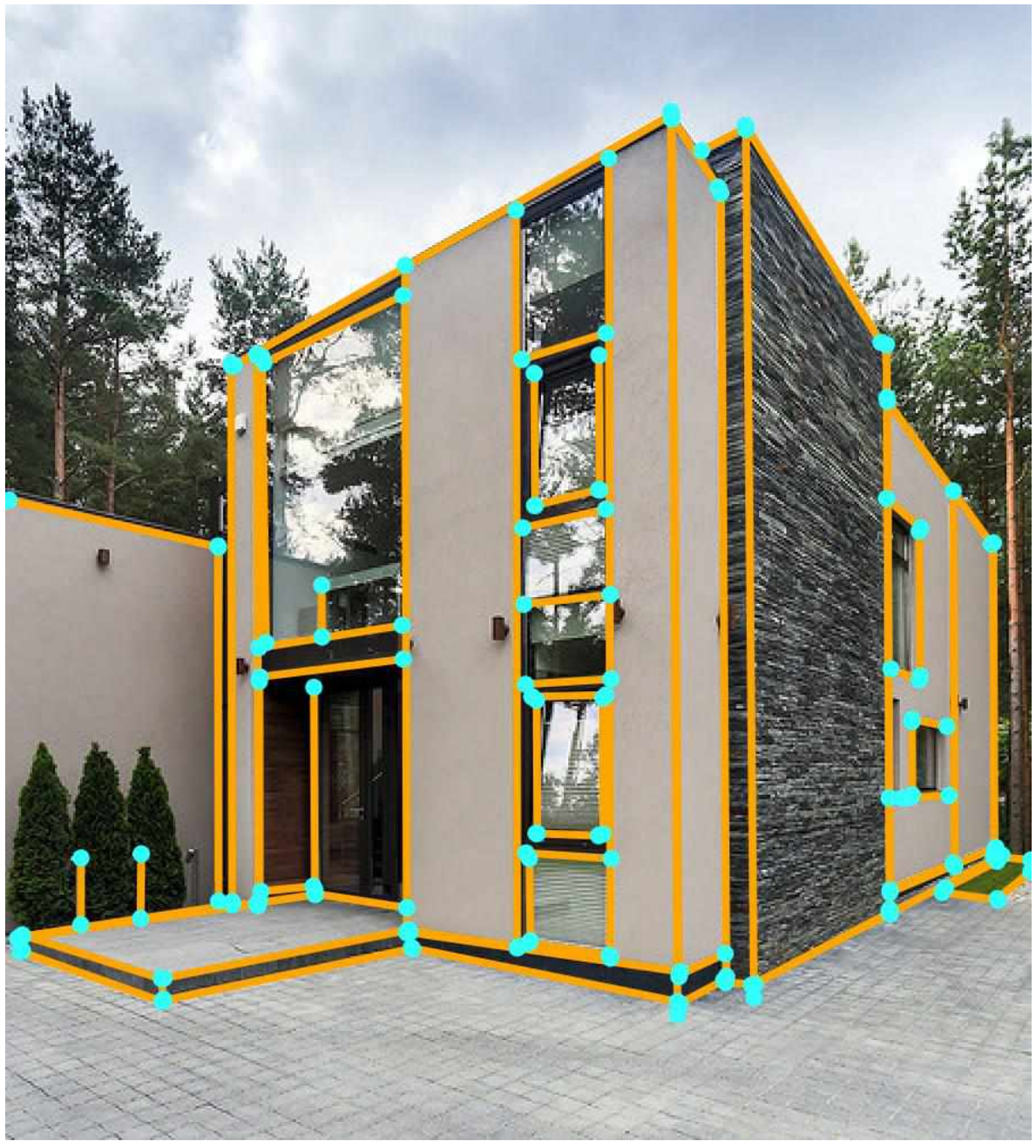}
    \includegraphics[width=0.99\linewidth]{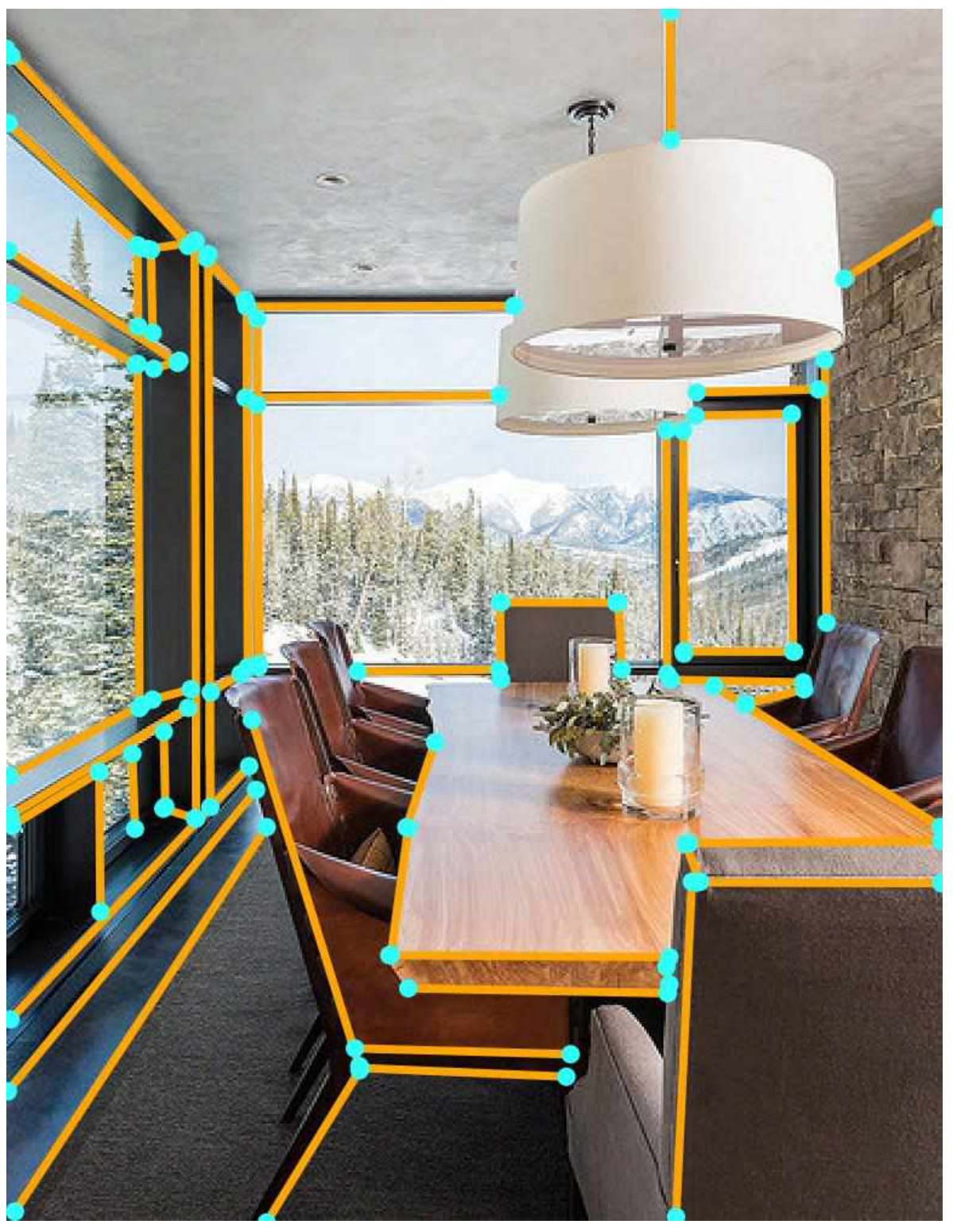}
    \includegraphics[width=0.99\linewidth]{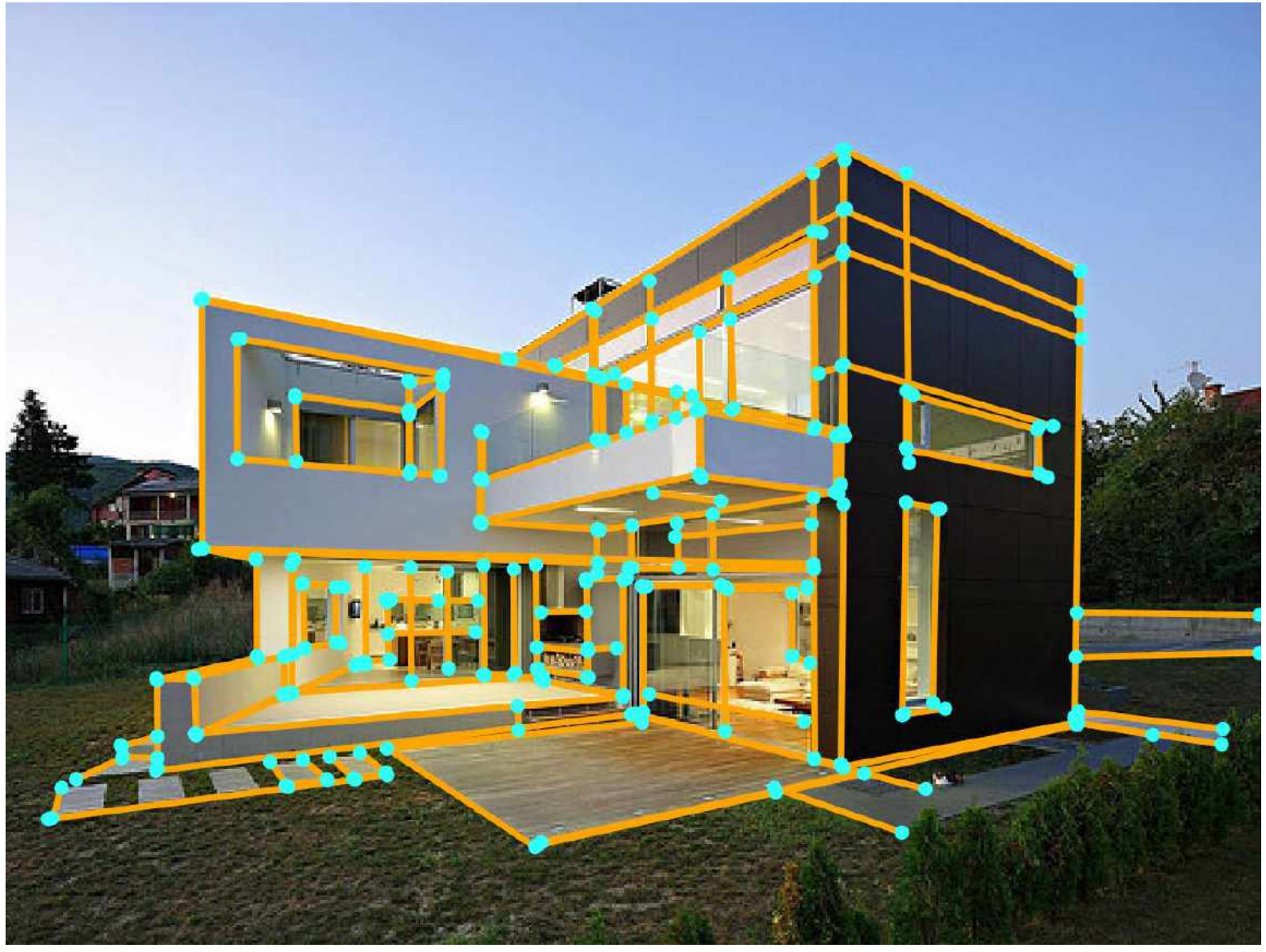}
    \includegraphics[width=0.99\linewidth]{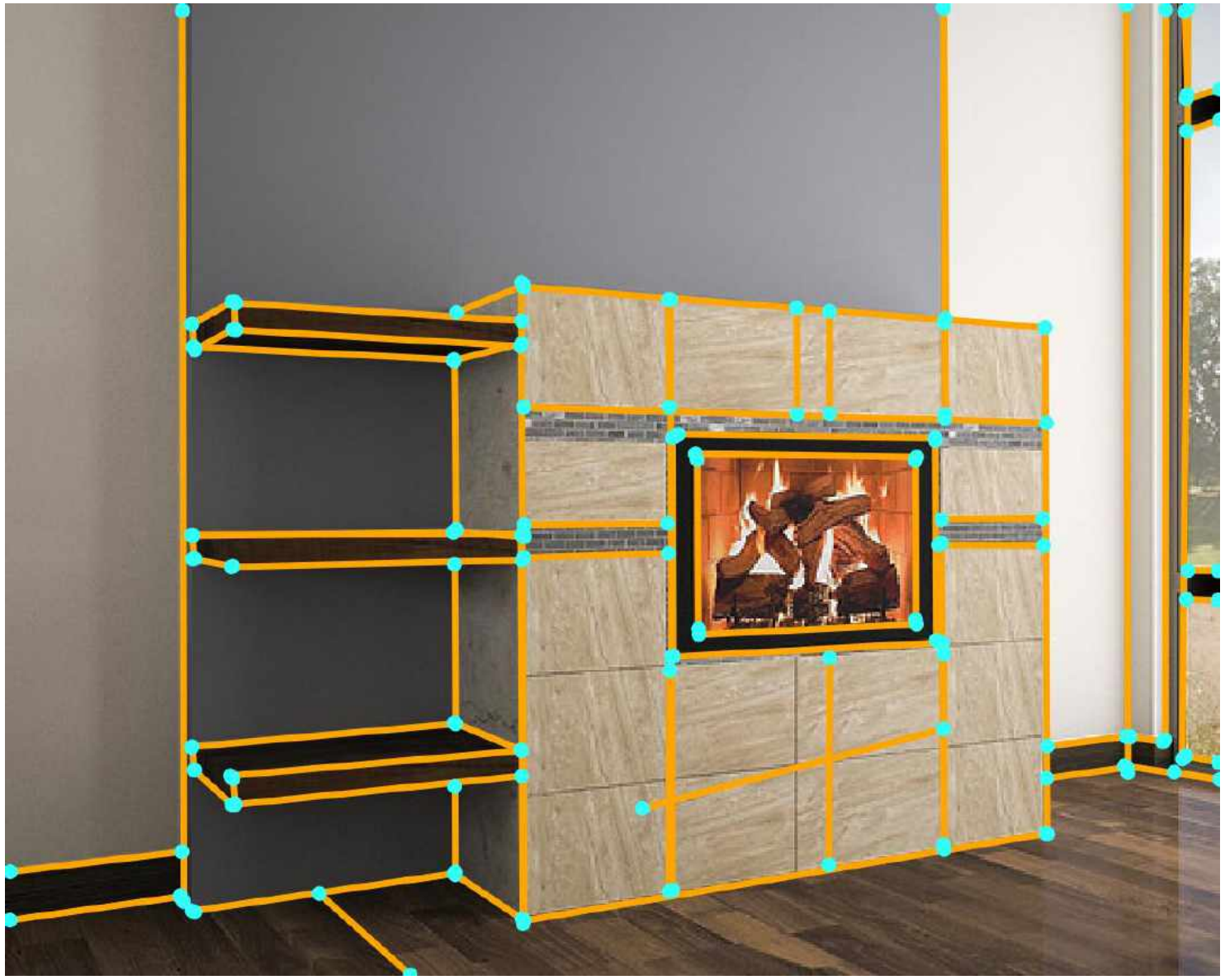}
    (c) TP-LSD
    \end{minipage}
    \begin{minipage}[t]{0.19\linewidth}\centering
    \includegraphics[width=0.99\linewidth]{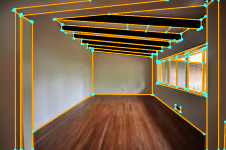}
    \includegraphics[width=0.99\linewidth]{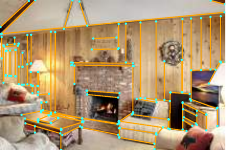}
    \includegraphics[width=0.99\linewidth]{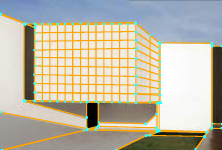}
    \includegraphics[width=0.99\linewidth]{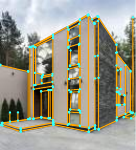}
    \includegraphics[width=0.99\linewidth]{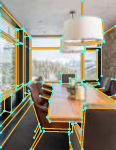}
    \includegraphics[width=0.99\linewidth]{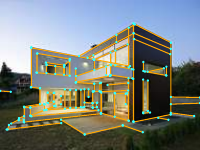}
    \includegraphics[width=0.99\linewidth]{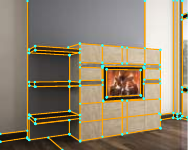}
    (d) \ours{} (HR)
    \end{minipage}
    \begin{minipage}[t]{0.19\linewidth}\centering
    \includegraphics[width=0.99\linewidth]{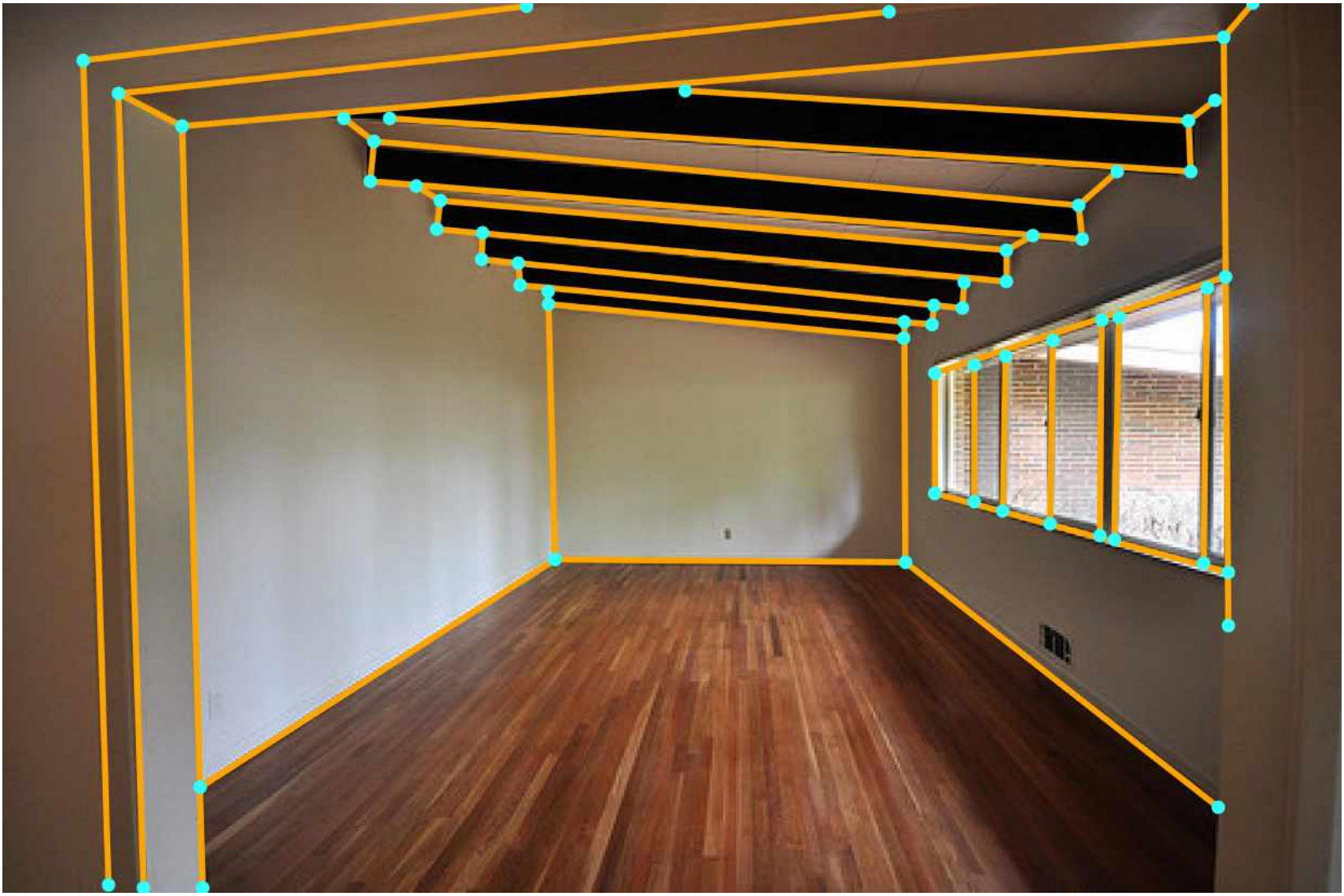}
    \includegraphics[width=0.99\linewidth]{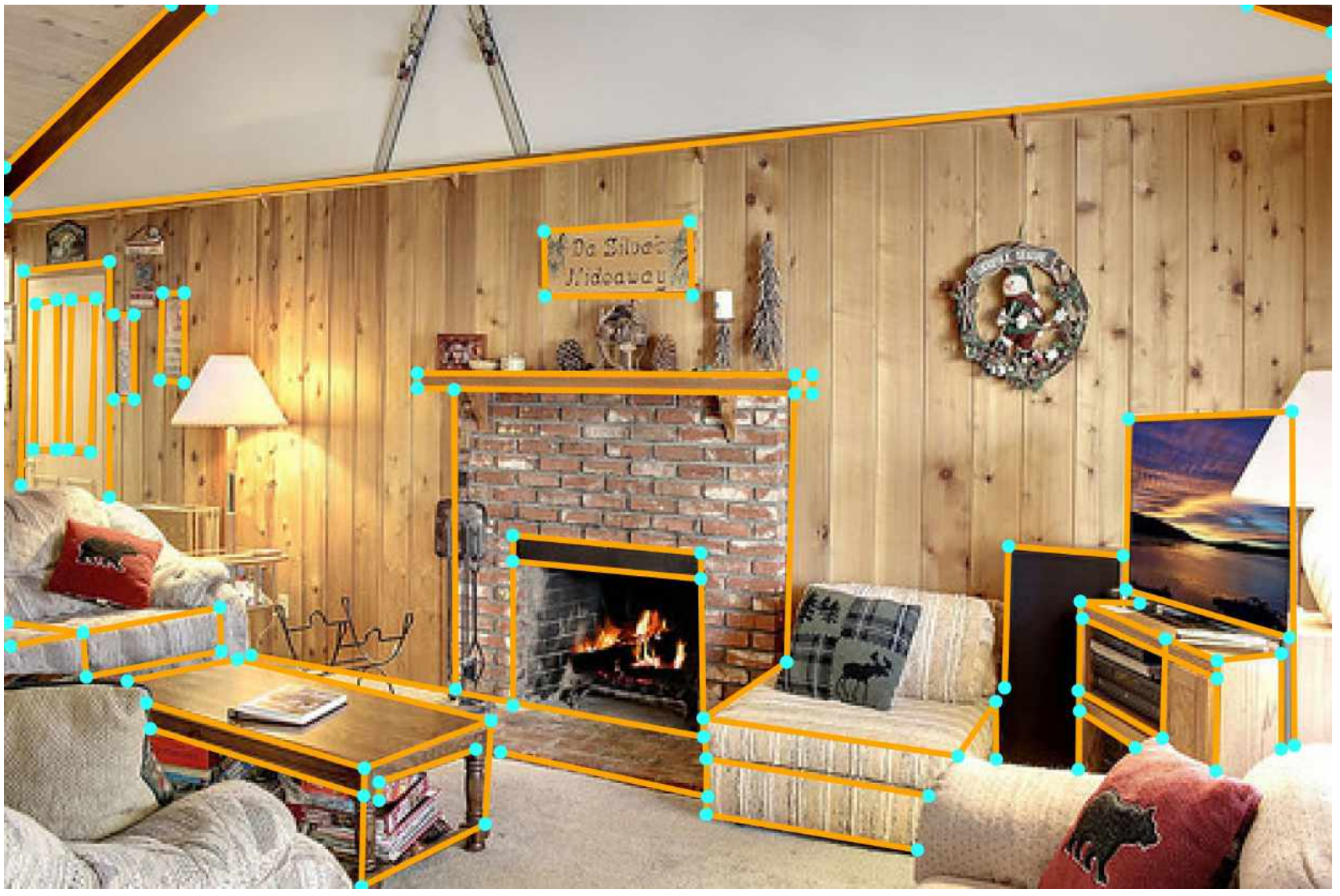}
    \includegraphics[width=0.99\linewidth]{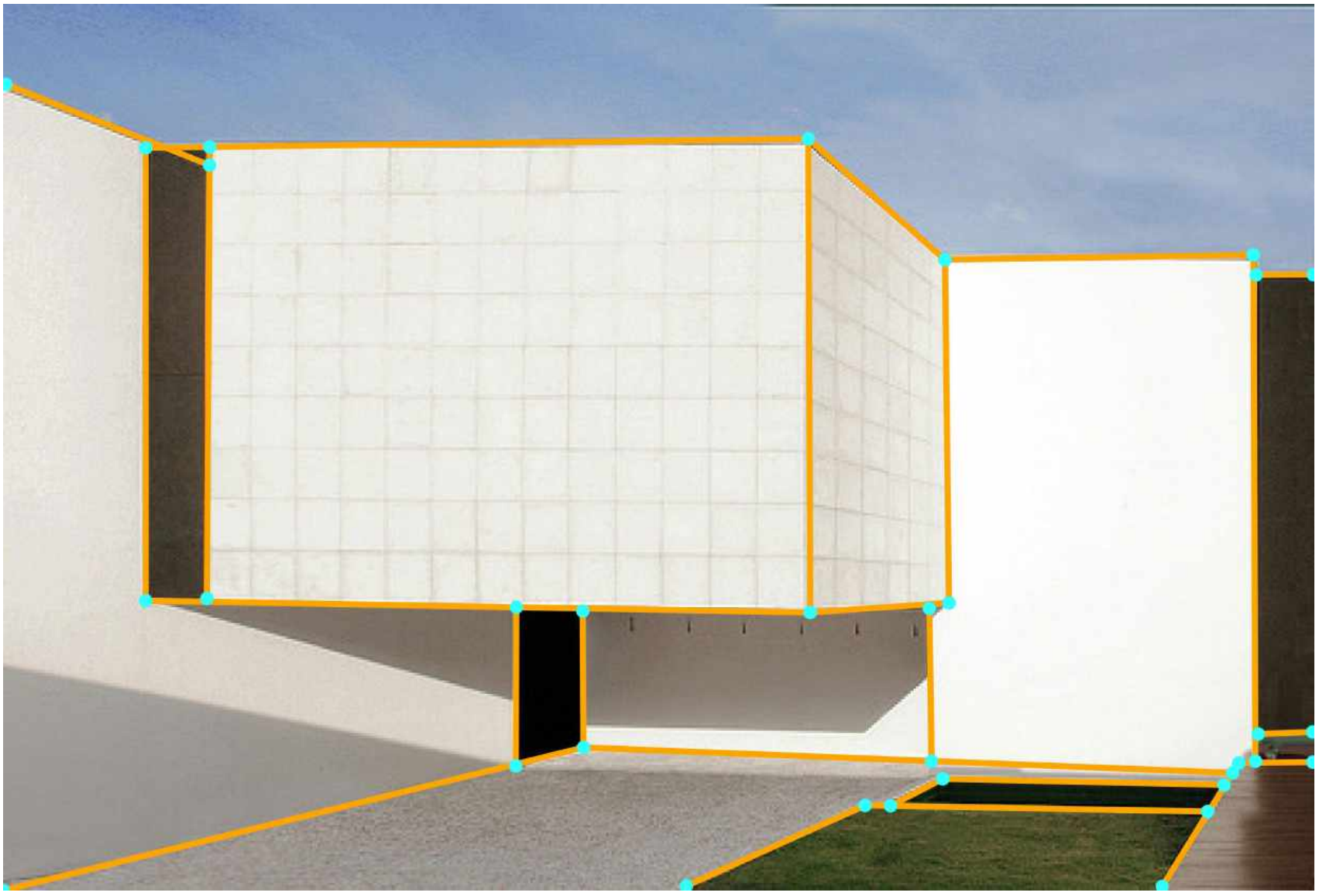}
    \includegraphics[width=0.99\linewidth]{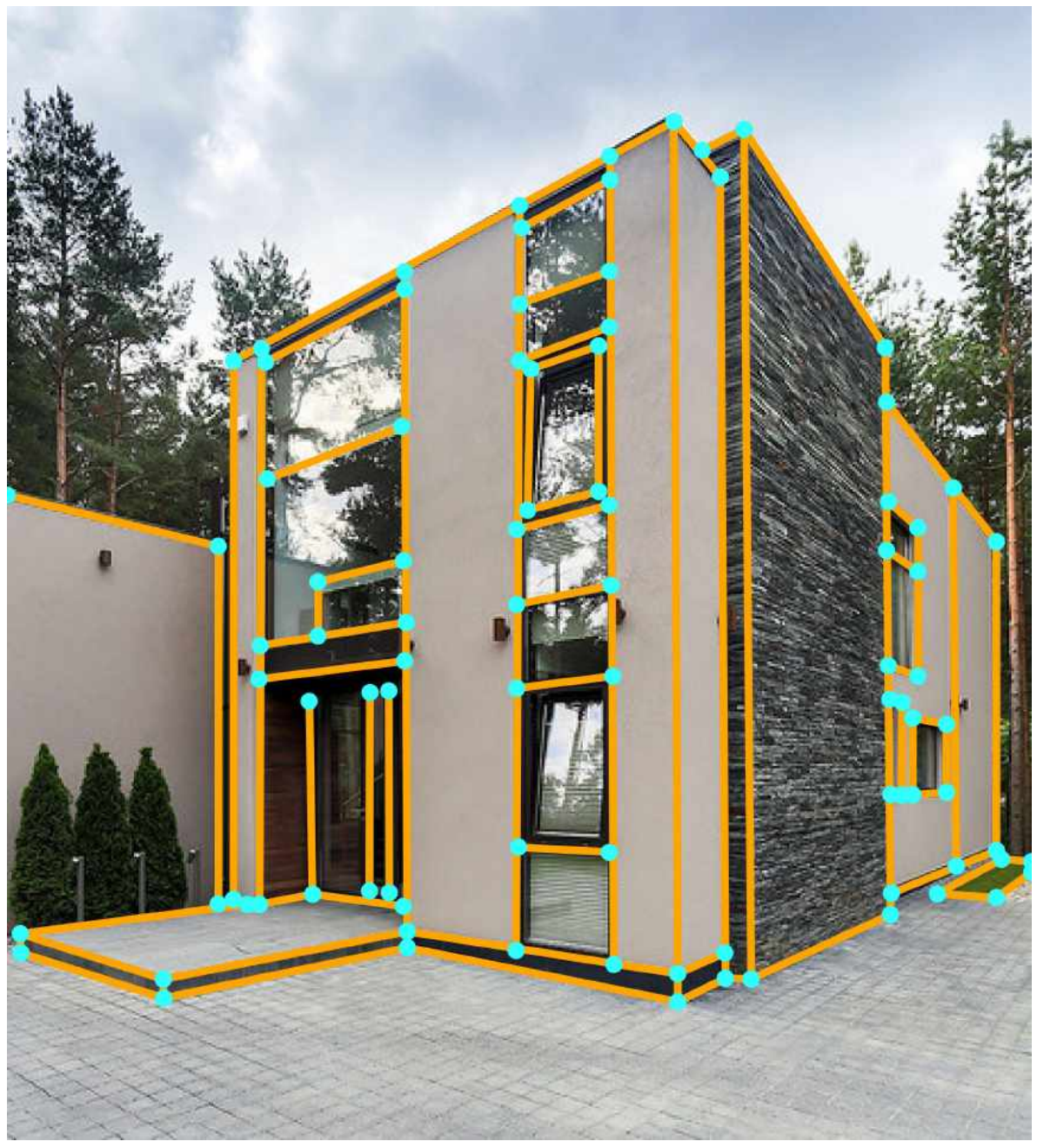}
    \includegraphics[width=0.99\linewidth]{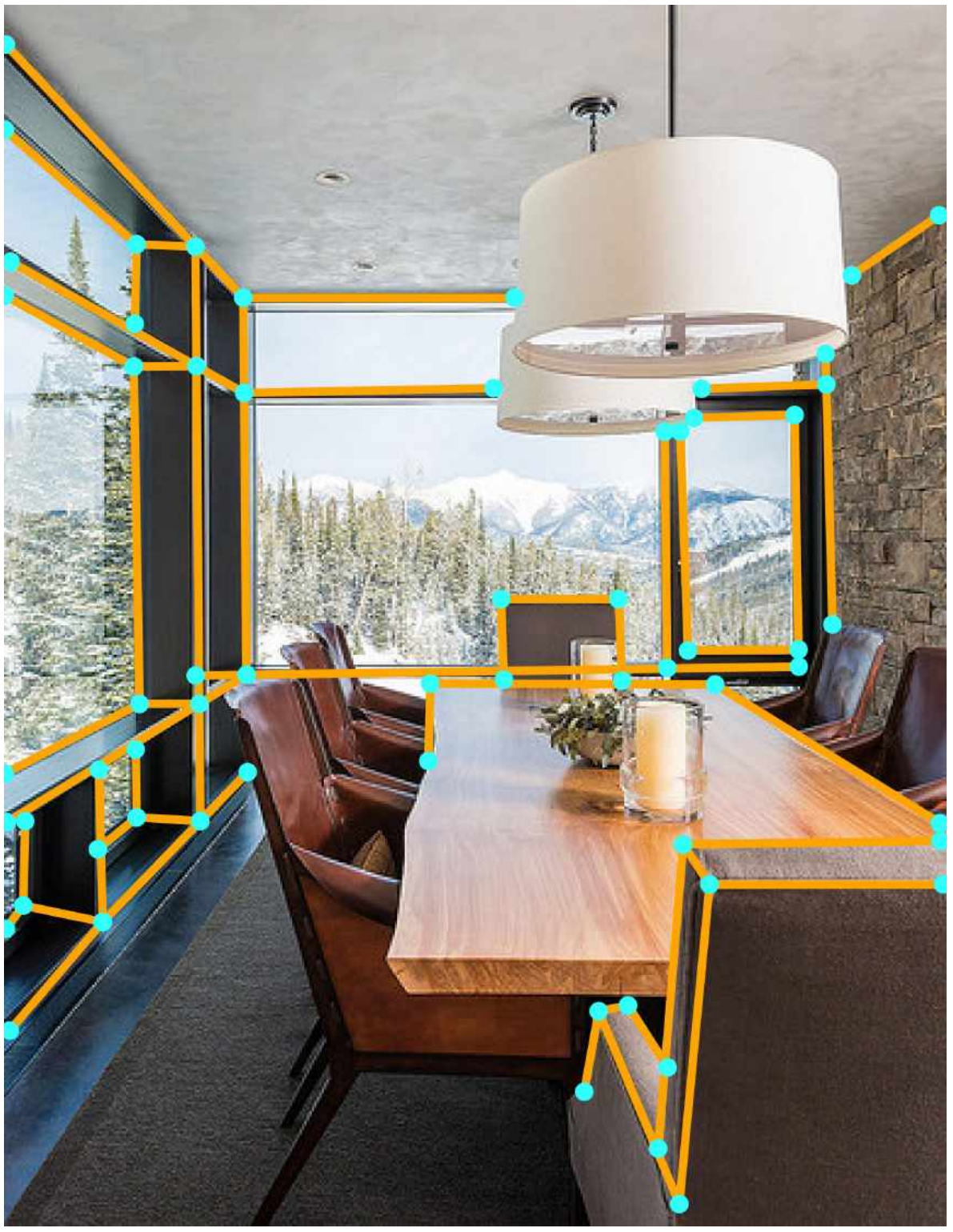}
    \includegraphics[width=0.99\linewidth]{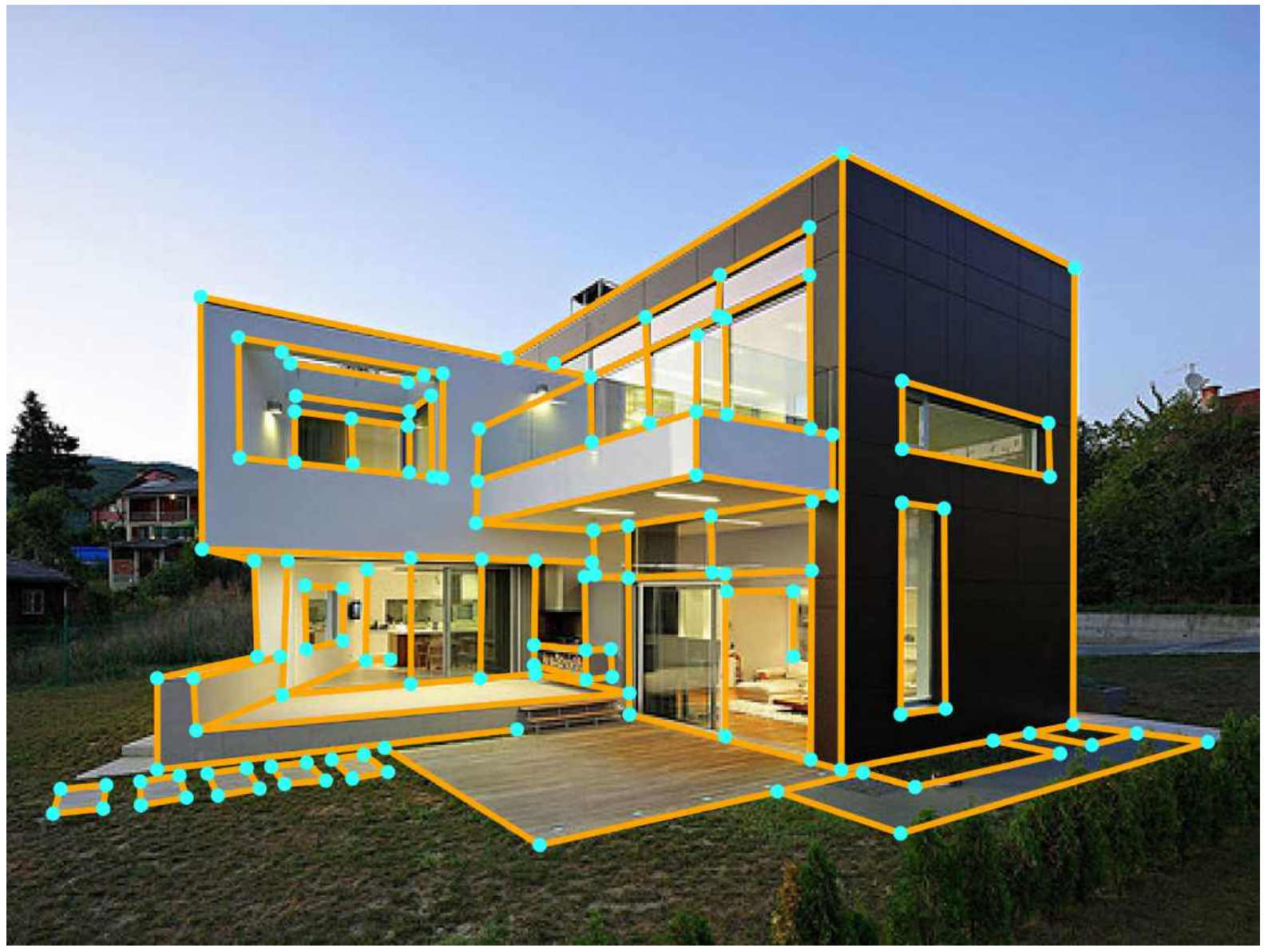}
    \includegraphics[width=0.99\linewidth]{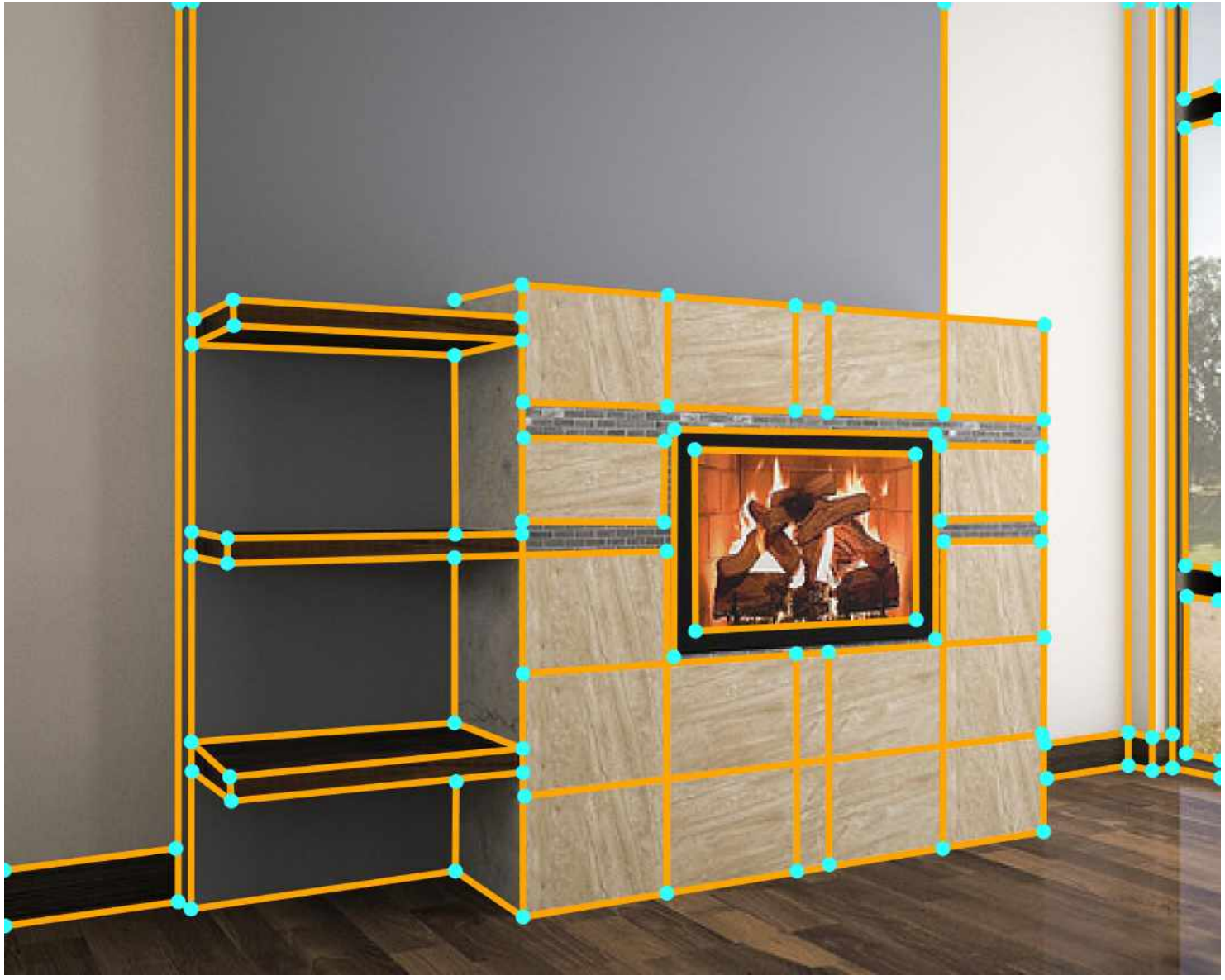}
    (e) GT
    \end{minipage}
    \caption{Qualitative evaluation of wireframe and line detection methods. From left to right, the columns correspond to the results from L-CNN~\cite{zhou2019end}, HAWP~\cite{xue2020holistically}, TP-LSD~\cite{huang2020tp}, \ours{} (HR), and the ground truth. We also draw the detected junctions from L-CNN and HAWP and the line endpoints from TP-LSD and \ours{}.}
    \label{fig:viz}
\end{figure*}

We visualize the output of our \ours{} and other three methods L-CNN, HAWP, and TP-LSD in Figure~\ref{fig:viz}. The junctions are marked cyan and lines are marked orange. Wireframes from L-CNN and HAWP are post processed using the method from Appendix A.1 in \cite{zhou2019end}. Since TP-LSD and~\ours{} do not explicitly output junctions, we treat the endpoints of lines as junctions. The last column GT means Ground Truth.

Both L-CNN and HAWP rely highly on the junction detection and line feature sampling, which might be prone to missing junctions or texture variations. In comparison, TP-LSD and and \ours{} are capable of detecting line segments in complicated even low-contrast environments (see third row of Figure~\ref{fig:viz}). An obvious draw-back of TP-LSD is that it captures many redundant lines (see details in Figure~\ref{fig:viz}).

\section{Conclusions}\label{sec:future}

In this work, we have introduced a one-stage fully convolutional line detection network \ours{} that directly outputs parameters of all line segments from an image. We formulate line segment detection as an end-to-end prediction of the center-point, length, and angle for each line segment. We show that by simply adjusting the backbone network, we are able to obtain a family of line detection networks that achieve state-of-the-art trade-off between accuracy and speed. We have conducted extensive experiments on large real-world datasets and demonstrated that this method outperforms the previous state-of-the-art wireframe parsing and line detection methods, by either improving the accuracy by a wide margin at the same frame rate or improving the speed by multi-fold at the same accuracy. 

\section{Acknowledgement}
We thank Yichao Zhou and Haozhi Qi of Berkeley for their help during all processes include ideas, experiments, and paper writing. This paper would not have been possible without their help. We also thank Kenji Tashiro of Sony for his helpful discussions.

{\small

\bibliographystyle{ieee_fullname}
}

\end{document}